\documentclass[]{xiaomi}
\usepackage{amsmath,amsfonts,amssymb}

\usepackage{colortbl}
\usepackage{makecell}
\usepackage{tabularx}
\usepackage{longtable}

\usepackage{fvextra}
\usepackage{adjustbox}
\usepackage{xspace}
\usepackage{pifont}
\usepackage{enumitem}

\definecolor{BrickRed}{rgb}{.72,0,0}
\definecolor{darkgreen}{rgb}{0.0,0.5,0.0}
\definecolor{ForestGreen}{RGB}{34,139,34}
\definecolor{LakeBlue}{RGB}{0,61,153}
\definecolor{MiOrange}{RGB}{255,105,0}
\definecolor{Orange}{RGB}{255,250,245}
\definecolor{Hex}{RGB}{225,213,231}
\definecolor{xiaomiorange}{RGB}{255,103,0}

\newcommand{\REPORTNAME}{Xiaomi-Lens-0\xspace}

\newcommand{\REPORTNAMEhl}{{\sffamily\bfseries\textcolor{xiaomiorange}{Xiaomi-Lens-0}}\xspace}
\newcommand{\takeaway}[2]{%
  \begin{tcolorbox}[enhanced, colback=Orange, colframe=xiaomiorange,
    boxrule=0.8pt, arc=4pt, left=4pt, right=4pt, top=3pt, bottom=3pt,
    before skip=8pt plus 2pt minus 2pt, after skip=8pt plus 2pt minus 2pt]
    \phantomsection\label{takeaway:#1}%
    \noindent\textbf{\textit{Takeaway #1:}} #2
  \end{tcolorbox}}

\titleformat{\paragraph}[runin]
  {\normalfont\normalsize\bfseries\color[HTML]{FF7E00}}
  {}{0pt}{}
\titlespacing*{\paragraph}{0pt}{6pt}{1em}


\title{\REPORTNAMEhl Technical Report}
\author{Xiaoai Plus, SeerRay Team}
\contribution{See \hyperref[contributors]{Contributors} for the full author list.}
\renewcommand\checkdataformat[2][]{{\small \ifstrempty{#1}{}{{\checkdatafont\sffamily\bfseries #1:} }#2}}
\checkdata{\vspace{6pt}\raisebox{-0.1ex}{\includegraphics[height=0.95em]{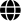}}~Homepage: \url{https://seerray-lab.github.io/Xiaomi-Lens-0/}}

\abstract{Multimodal large language models (MLLMs) answer questions about whole images well, yet they still struggle with fine-grained region understanding. A key reason is that text-only supervision never specifies which regions support an answer. To close this gap, we present \REPORTNAMEhl, a 4B MLLM built on Qwen3.5 that predicts masks from its own visual representations. Specifically, a lightweight decoder combines vision-encoder patch features with the language model's mask-token states, termed visual latents, so that mask supervision reaches the shared backbone without an additional segmentation encoder. Because each latent decodes to a mask, the model can interleave reasoning with region-specific latents and thus think with visual latents, while grounded reasoning traces supervise the intermediate regions referenced during chain-of-thought. As a result, across all eight RefCOCO, RefCOCO+, and RefCOCOg splits, \REPORTNAME gains 5.6 points on average over Qwen3.5 in referring comprehension and achieves the highest segmentation cIoU among compared models. Moreover, it surpasses Qwen3.5 across all perception benchmarks.
}

\begin{document}
\maketitle

\begin{center}
    \includegraphics[width=\textwidth]{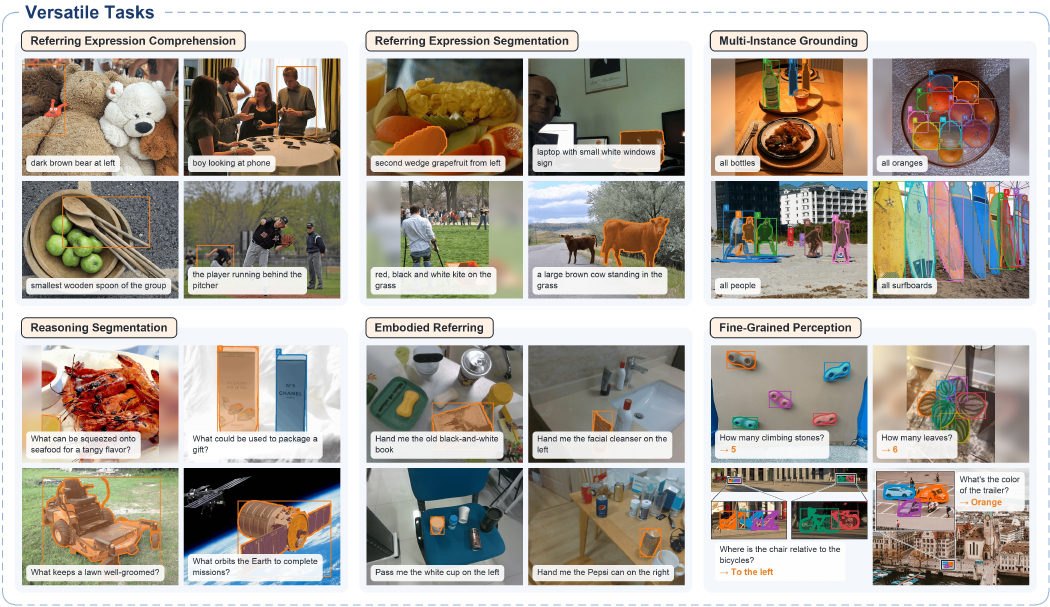}
    \captionsetup{skip=4pt}
    \captionof{figure}{Examples of \REPORTNAME across six tasks: referring expression comprehension, referring expression segmentation, multi-instance grounding, reasoning segmentation, embodied referring, and fine-grained perception. The model localizes targets with boxes or masks and answers questions about counting, attributes, and spatial relations.}
    \label{fig:versatile-tasks}
\end{center}

\clearpage
\tableofcontents

\newpage
\section{Introduction}
\label{sec:introduction}

Visual grounding, which links natural-language expressions to image regions, is a foundational capability of multimodal large language models (MLLMs)~\citep{bai2025qwen3,zhu2025internvl3}, supporting embodied intelligence~\citep{lu2023vlgrasp} and multimodal agent systems~\citep{cheng2024seeclick,gou2025uground}. Specifically, referring expression comprehension (REC) and referring expression segmentation (RES) evaluate grounding directly, requiring the model to output the referenced target as a bounding box or a pixel-level mask, respectively. Beyond these tasks, visual questions about object counting, attributes, and spatial relations also depend on grounding: answering them requires pinpointing the referenced regions and perceiving their visual details~\citep{wu2023vstar,wang2024divide}.

However, current MLLMs still struggle with fine-grained region understanding, often miscounting nearby objects or misidentifying the attributes of small regions~\citep{tamarapalli2025countqa,liang2025colorbench}. We attribute this to text-only supervision, under which the image serves only as a conditioning input and the training signal comes entirely from the text answer. Such supervision does not explicitly specify which image regions support the answer or how far those regions extend. By contrast, segmentation masks provide explicit spatial targets by labeling which pixels belong to a referenced region~\citep{kirillov2023segment}, making them a direct source of supervision for region extent.

Although existing MLLMs for RES~\citep{xia2024gsva,rasheed2024glamm,zhang2024omgllava} already train with mask supervision, most obtain the spatial detail for mask prediction from an additional encoder: LISA~\citep{lai2024lisa} and Sa2VA~\citep{yuan2025sa2va} rely on a hierarchical segmentation encoder~\citep{kirillov2023segment,ravi2024sam2}, and Qwen3-VL-Seg~\citep{yao2026qwen3vlseg} fuses multi-scale features from a lightweight supplementary image encoder. Others decode masks from the MLLM's own visual features: PixelLM~\citep{ren2024pixellm} decodes from multiple layers of its vision encoder, and MLLMSeg~\citep{wang2025mllmseg} fuses shallow vision-encoder features with the language model's visual tokens. However, both keep the vision encoder frozen, so the mask loss does not reach the shared visual features. Moreover, across both lines, mask supervision is treated as a means to segmentation and evaluated mainly on RES and grounding, leaving unexplored whether it can also improve fine-grained perception and visual reasoning. Therefore, we investigate \textbf{whether dense mask supervision can improve fine-grained perception and visual reasoning by requiring the MLLM to predict masks from its own visual representations}.

In this report, we develop \REPORTNAME, a 4B MLLM built on Qwen3.5~\citep{qwen3.5} that predicts region masks directly from its own visual representations, bringing pixel-level supervision into the backbone shared with language generation. Concretely, the model reads spatial features from the patch grid of its vision encoder~\citep{dosovitskiy2021image}, taken before the spatial merger, without an additional hierarchical visual encoder or multi-scale feature fusion. To specify the target region, it generates a special mask token~\citep{lai2024lisa} whose contextual state, termed a visual latent, represents the region requested by the instruction or referenced in the preceding response. A lightweight mask decoder combines this visual latent with the shared visual features to predict the corresponding mask. Joint language and mask training updates both the shared visual features and the language model states that identify target regions, thereby giving the representations used for perception and language generation explicit supervision over region extent.

In this architecture, the mask decoder also serves as a visual latent decoder. This allows the MLLM to interleave textual reasoning with region-specific visual latents~\citep{yang2025mirage,li2026lvr}, each of which is decoded into a mask that explicitly grounds the corresponding reasoning evidence in the image. We refer to this formulation as \textbf{thinking with visual latents}. To train this capability, we construct grounded reasoning traces~\citep{shao2024visualcot,liu2025unipixel} that pair intermediate region references with masks, and jointly supervise both the textual responses and the decoded masks. This extends pixel-level supervision to intermediate regions referenced in chain-of-thought (CoT)~\citep{wei2022chain}, making their spatial extent explicit and inspectable through decoded masks.

We evaluate \REPORTNAME across six tasks (Figure \ref{fig:versatile-tasks}): REC, RES, multi-instance grounding, reasoning segmentation, embodied referring, and fine-grained perception. It improves REC over its Qwen3.5 backbone on all eight splits of RefCOCO, RefCOCO+~\citep{yu2016modeling}, and RefCOCOg~\citep{mao2016generation,nagaraja2016modeling}, with an average gain of 5.6 points, and achieves the highest cumulative intersection over union (cIoU) among the compared models on every RES split. For multi-instance grounding on COCO 2017~\citep{lin2014coco}, it also achieves the highest F1 averaged over intersection over union (IoU) thresholds from 0.50 to 0.95. Beyond spatial prediction, it outperforms its backbone on every perception benchmark. To evaluate referring under conditions that existing benchmarks rarely cover, we also build LensBench, which crosses explicit and implicit instructions with single and multiple targets across general, aerial, driving, and panoramic imagery, with every paired unit human-reviewed for target completeness and instruction--mask agreement (Section \ref{sec:lensbench}). \REPORTNAME achieves the highest cIoU among the compared models in the aerial, driving, and panoramic domains, although implicit instructions remain challenging, particularly when selecting multiple targets in panoramic scenes. Finally, ablations show that grounded reasoning traces further increase REC and RES scores.

Our main contributions can be summarized as follows:
\begin{itemize}[leftmargin=1.5em]
    \item \textbf{Mask prediction from shared visual representations.} We develop \REPORTNAME, a 4B MLLM that decodes region masks from the patch grid of its own vision encoder, taken before the spatial merger, and the visual latents produced by its language model, without an additional hierarchical vision encoder or multi-scale feature fusion. Joint language and mask training therefore supervises region extent directly in the representations shared with language generation.

    \item \textbf{Thinking with visual latents.} We interleave textual reasoning with visual latents that the mask decoder maps to masks, and construct grounded reasoning traces that extend pixel-level supervision to intermediate regions in CoT.

    \item \textbf{LensBench.} We introduce a human-reviewed referring benchmark that pairs single and multiple targets with explicit and implicit instructions across four visual domains, enabling paired comparisons of target selection and mask prediction under varying instruction semantics.

    \item \textbf{Mask supervision also improves perception.} Beyond the spatial gains above, \REPORTNAME outperforms its Qwen3.5 backbone on every fine-grained perception benchmark. Moreover, ablations show that adding mask supervision to box supervision further improves all three evaluated general perception benchmarks and the REC validation splits of RefCOCO, RefCOCO+, and RefCOCOg.
\end{itemize}

\FloatBarrier
\section{Method}
\label{sec:method}

\begin{figure}[!tp]
\centering
\includegraphics[width=\textwidth]{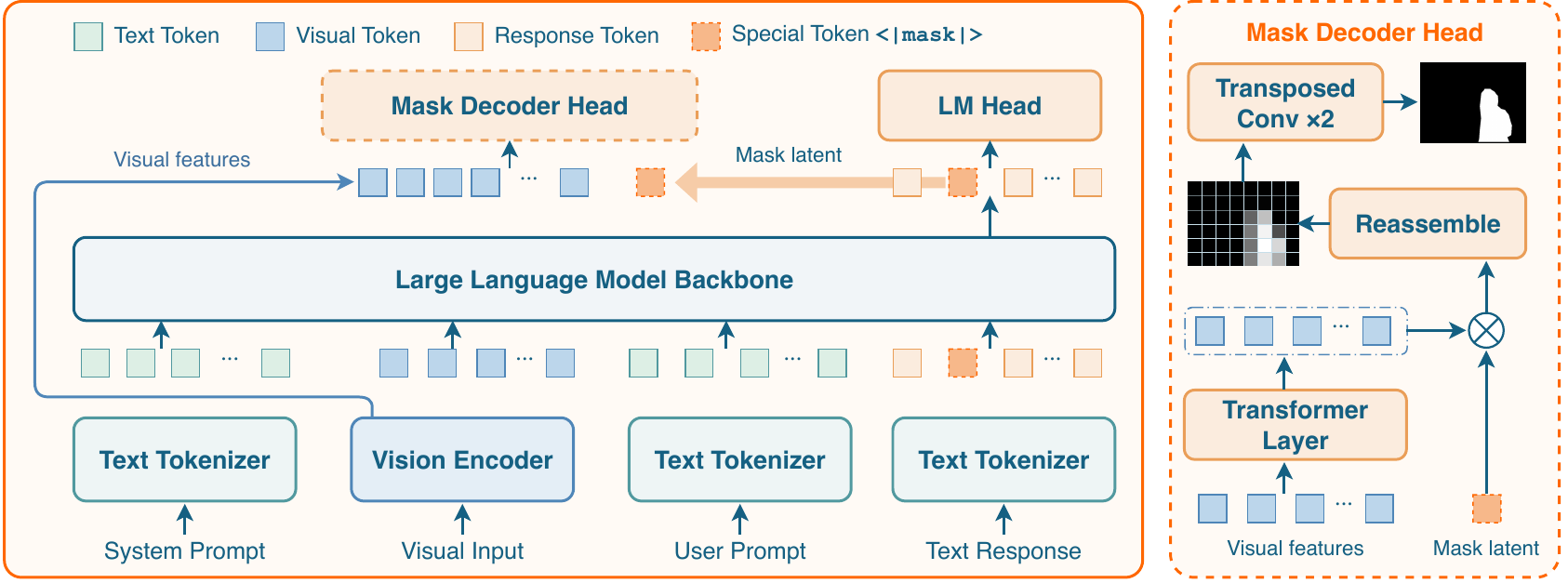}
\caption{\REPORTNAME architecture. The language model consumes spatially merged visual tokens. Each generated \texttt{\textless{}|mask|\textgreater{}} token has a contextual state that queries the encoder's patch-level features through a lightweight mask decoder. Patch-wise compatibility scores are reassembled into a spatial grid and upsampled with two transposed-convolution layers.}
\label{fig:method}
\end{figure}

\REPORTNAME predicts masks directly from its own visual representations, so mask supervision must reach the same features and states that support language generation. As shown in Figure \ref{fig:method}, the shared vision encoder supplies merged tokens to the language model and its patch-level feature grid to a lightweight mask decoder. The decoder combines this grid with a region-specific visual latent from the language model, and joint training propagates the mask loss into both. We first motivate decoding from the shared encoder (Section \ref{sec:rethinking-dense}), then describe the architecture and joint objective (Section \ref{sec:model-architecture}), extend mask prediction to intermediate regions through thinking with visual latents (Section \ref{sec:thinking-visual-latents}), and present the two-stage training recipe (Section \ref{sec:training-recipe}).

\subsection{Mask Prediction from Shared Visual Features}
\label{sec:rethinking-dense}

Mask prediction requires spatial features that describe region extent and a query that identifies the target. In LISA and Sa2VA, the spatial features come from an additional hierarchical segmentation encoder, and Qwen3-VL-Seg fuses multi-scale features from a supplementary image encoder. Under these designs, the MLLM's own visual representation is not required to encode the region detail that the mask loss supervises. To supervise the shared representation directly, we decode masks from a single patch grid obtained before the MLLM's spatial merger, without an additional encoder or multi-scale fusion. This choice preserves the encoder's patch resolution and provides the mask loss with a direct path to the visual features used for language generation.

The patch grid of a Vision Transformer (ViT)~\citep{dosovitskiy2021image} provides spatially indexed features enriched by attention across the image. Its residual structure expresses the final representation as the initial patch features plus successive transformations:
\begin{equation}
    \mathbf{x}_{l}=\mathbf{x}_{l-1}+\mathcal{F}_{l}(\mathbf{x}_{l-1}),
    \qquad
    \mathbf{x}_{L}=\mathbf{x}_{0}+\sum_{l=1}^{L}\mathcal{F}_{l}(\mathbf{x}_{l-1}),
    \label{eq:vit-residual-recovery}
\end{equation}
where $\mathbf{x}_0$ contains the initial patch features and $\mathcal{F}_l$ is the residual transformation at layer $l$. The identity mapping $\mathbf{x}_{l-1}$ keeps the fine-grained signals produced by earlier layers from being overwritten by higher-level semantic abstractions. Instead, those signals propagate through the network and accumulate into $\mathbf{x}_{L}$, so the final representation retains a high-mutual-information pathway to the input image.

LongCat-Next formalizes this property as intrinsic information recovery and tests it by pairing frozen encoders with a lightweight pixel decoder~\citep{longcatnext2026}. Specifically, semantic encoders with residual pathways reconstruct coarse image structure despite receiving no reconstruction supervision, and a randomly initialized ViT even reconstructs better than its pretrained counterpart, which indicates that recoverability follows from the network structure rather than from learned semantics. Moreover, QwenViT~\citep{bai2025qwen3}, without its spatial merger, behaves like a plain ViT that maintains a single patch resolution and still recovers coarse image structure, whereas the merger degrades fidelity because its aggressive downsampling discards fine-grained, high-frequency detail. Because the patch grid is taken before the merger, it preserves the spatial structure required for mask prediction while remaining part of the encoder's shared computation. We therefore feed this patch-level grid to the mask decoder and query it with region-specific contextual states from the language model, allowing a single shared grid to support multiple mask predictions.

\subsection{Model Architecture}
\label{sec:model-architecture}

\subsubsection{Shared Visual Features and Visual Latents}
Language generation and mask prediction begin with the same image encoding. Given an image $I\in\mathbb{R}^{3\times H\times W}$, the vision encoder $\mathcal{E}_v$ produces a spatial grid of patch features, which we flatten for notational convenience:
\begin{equation}
    \mathbf{V}^{0}=\mathcal{E}_{v}(I)
    = [\mathbf{v}_{1},\ldots,\mathbf{v}_{N_p}]
    \in\mathbb{R}^{N_p\times d_v},
    \qquad N_p=h_vw_v,
    \label{eq:patch-features}
\end{equation}
where $(h_v,w_v)$ denotes the patch-grid resolution. Following the Qwen3.5 backbone~\citep{qwen3.5}, an $s\times s$ spatial merger groups neighboring patch features, and the visual projector $\phi$ maps the merged features to the language-model width $d_l$:
\begin{equation}
    \mathbf{V}=\phi\!\left(\operatorname{Merge}_{s}(\mathbf{V}^{0})\right)
    \in\mathbb{R}^{N_v\times d_l},
    \qquad N_v=N_p/s^2.
    \label{eq:merged-features}
\end{equation}
The autoregressive language model processes the merged tokens $\mathbf{V}$ together with the system and user tokens. At decoding step $t$, the language-model head maps the contextual state $\mathbf{h}_t\in\mathbb{R}^{d_l}$ to a next-token prediction.

To mark each region for mask decoding, we add a special \texttt{\textless{}|mask|\textgreater{}} token to the output vocabulary. For its $k$-th occurrence at response position $t_k$, we take the contextual state $\mathbf{m}_k=\mathbf{h}_{t_k}\in\mathbb{R}^{d_l}$ as the visual latent. The token identity is shared across occurrences, but each state depends on the image, the instruction, and the preceding response, so it represents the region requested by the instruction or referenced in the preceding response. A response can contain several mask tokens, each supplying the query for a separate mask.

\subsubsection{Mask Decoding}
The mask decoder scores each patch of the grid $\mathbf{V}^{0}$ (Equation \ref{eq:patch-features}) for the requested target. A lightweight Transformer layer first refines these features with global visual context, and a linear projection then maps each patch to a mask feature space of width $d_m$:
\begin{equation}
    \mathbf{F}=P_v\!\left(\mathcal{T}_{\mathrm{mask}}(\mathbf{V}^{0})\right)
    =[\mathbf{f}_1,\ldots,\mathbf{f}_{N_p}]
    \in\mathbb{R}^{N_p\times d_m}.
    \label{eq:dense-visual-features}
\end{equation}
Next, we project the mask-token state into the same space, $\mathbf{q}_k=P_m(\mathbf{m}_k)\in\mathbb{R}^{d_m}$, and use it as a dynamic classifier over patch locations. Its compatibility with patch $p$ is
\begin{equation}
    z_{k,p}=\frac{\mathbf{q}_k^{\mathsf T}\mathbf{f}_p}{\sqrt{d_m}},
    \qquad p=1,\ldots,N_p.
    \label{eq:mask-compatibility}
\end{equation}
Finally, we reassemble the $N_p$ scores into the original patch layout and apply two transposed convolution layers to produce a dense mask logit map. Applying a sigmoid gives
\begin{equation}
    \widehat{\mathbf{M}}_k
    =\sigma\!\left(
        \mathcal{U}_{\mathrm{mask}}
        \bigl(\operatorname{Reassemble}_{h_v,w_v}(\mathbf{z}_k)\bigr)
      \right)
    \in[0,1]^{H\times W},
    \label{eq:mask-prediction}
\end{equation}
where $\mathcal{U}_{\mathrm{mask}}$ is the two-layer upsampling head and $\sigma$ is the sigmoid function. Because all regions share the same image features, we compute $\mathbf{F}$ once per image and reuse it for every visual latent, changing only the query $\mathbf{q}_k$. Beyond the shared vision encoder, the decoder consists of only one Transformer refinement layer, two linear projections, and a two-layer upsampling head.

\subsubsection{Joint Language and Mask Supervision}
Pixel-level mask supervision gives each visual latent an explicit target region. For a training example with $K$ supervised \texttt{\textless{}|mask|\textgreater{}} occurrences, the $k$-th occurrence is paired with a binary target mask $\mathbf{M}_k$. The mask objective combines pixel-wise binary cross-entropy and Dice loss~\citep{milletari2016vnet}:
\begin{equation}
    \mathcal{L}_{\mathrm{mask}}
    =\frac{1}{K}\sum_{k=1}^{K}
      \left[
        \lambda_{\mathrm{bce}}\,
        \mathcal{L}_{\mathrm{bce}}(\widehat{\mathbf{M}}_k,\mathbf{M}_k)
        +
        \lambda_{\mathrm{dice}}\,
        \mathcal{L}_{\mathrm{dice}}(\widehat{\mathbf{M}}_k,\mathbf{M}_k)
      \right],
    \label{eq:mask-loss}
\end{equation}
where $\lambda_{\mathrm{bce}}$ and $\lambda_{\mathrm{dice}}$ weight the two terms. The cross-entropy term supervises individual pixels, and the Dice term supervises region overlap. Adding the mask objective to the next-token prediction loss gives the joint objective:
\begin{equation}
    \mathcal{L}=\mathcal{L}_{\mathrm{ntp}}
    +\mathcal{L}_{\mathrm{mask}}.
    \label{eq:joint-loss}
\end{equation}
During joint training, the mask loss is propagated to the vision encoder via $\mathbf{F}$ and to the language model via $\mathbf{m}_k$, while $\mathcal{L}_{\mathrm{ntp}}$ supervises the response sequence. Region extent must therefore be encoded in the shared image features and in the contextual states that identify targets. Since text answers are generated from these same representations, mask supervision also improves fine-grained perception, as Section \ref{sec:ablation} shows.

\subsection{Thinking with Visual Latents}
\label{sec:thinking-visual-latents}

Because each visual latent is a contextual state of the language model, the mask decoder also decodes visual latents produced during reasoning. Specifically, in thinking with visual latents, the response interleaves textual reasoning with \texttt{\textless{}|mask|\textgreater{}} tokens placed at region references. Each token's contextual state represents the region discussed in the preceding text, and the decoder maps that state to a mask over the shared image grid. A counting trace, for example, can enumerate individual instances with one visual latent each, and a spatial comparison can assign separate latents to the objects being compared. Each latent is tied to an explicit image region.

We supervise these intermediate regions using grounded reasoning traces that pair region references with target masks (Section \ref{sec:reasoning-data}). The next-token loss supervises the reasoning text, spatial references, and final answer, while the mask loss supervises the extent of each annotated region. The joint objective in Equation \ref{eq:joint-loss} supervises both final segmentation targets and intermediate regions in a CoT.

During inference, mask decoding is an output branch alongside autoregressive generation: subsequent text is conditioned on the image and the preceding response tokens, and no crops or region features are fed back to the language model. Thus, the decoded masks make the regions behind intermediate statements inspectable, while training extends pixel-level supervision to the states that represent those regions.

\subsection{Training Recipe}
\label{sec:training-recipe}

\begin{figure}[!tp]
\centering
\includegraphics[width=\textwidth]{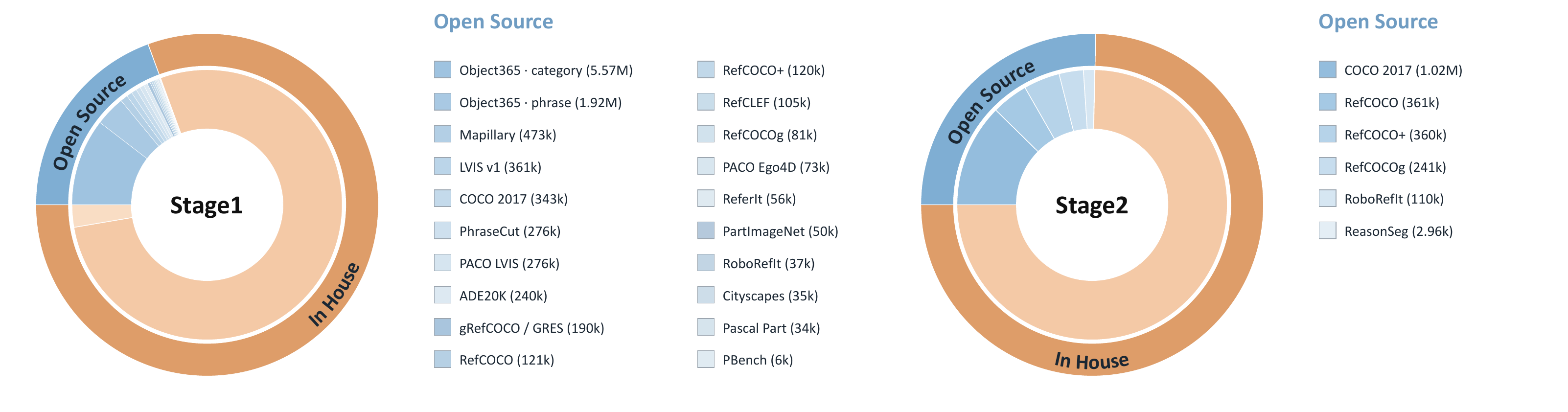}
\caption{Training data for Stage 1 mask decoder pretraining (left) and Stage 2 supervised fine-tuning (right). Donut charts show the proportions of public and in-house data; the adjacent lists give the public datasets and their sample counts.}
\label{fig:training-recipe}
\end{figure}

We train in two stages. Stage 1 pretrains the mask decoder to map frozen representations to regions, and Stage 2 jointly fine-tunes the backbone and decoder. Figure \ref{fig:training-recipe} summarizes the data mixtures. The in-house data, detailed in Section \ref{sec:data-pipeline}, provide masks for referring-expression targets and for intermediate regions in grounded reasoning traces.

\subsubsection{Stage 1: Mask Decoder Pretraining}
We freeze the pretrained backbone and optimize only the mask decoder with Equation \ref{eq:mask-loss}, so that it learns to decode regions from fixed image features and visual latents. The training mixture combines in-house data with public sources covering objects, parts, and referring expressions, including Objects365~\citep{shao2019objects365}, LVIS~\citep{gupta2019lvis}, PACO~\citep{ramanathan2023paco}, PhraseCut~\citep{wu2020phrasecut}, RefCOCO, RefCOCO+, and RefCOCOg. These sources expose the decoder to regions at different scales and levels of specificity before joint adaptation.

\subsubsection{Stage 2: Supervised Fine-Tuning}
We then unfreeze the backbone and optimize the full model with Equation \ref{eq:joint-loss}. The language loss supervises the response sequences, and the mask loss updates both the shared visual features and the visual latents. The data combine in-house referring-expression and grounded reasoning examples with COCO 2017, RefCOCO, RefCOCO+, RefCOCOg, RoboRefIt~\citep{lu2023vlgrasp}, and ReasonSeg~\citep{lai2024lisa}.

\FloatBarrier
\section{Data Pipeline}
\label{sec:data-pipeline}

The data pipeline supplies the two forms of region supervision used in Section \ref{sec:method}. Specifically, referring-expression data pair instructions with final target masks to supervise REC and RES outputs (Section \ref{sec:res-data}), whereas grounded reasoning data attach masks to the regions referenced at intermediate steps of a CoT to supervise thinking with visual latents (Section \ref{sec:reasoning-data}). Both pipelines fix or verify the region annotation before accepting the accompanying language. We then present LensBench (Section \ref{sec:lensbench}), which evaluates target selection and region extent as instruction semantics, target cardinality, and visual domain vary.

\subsection{Referring-Expression Training Data}
\label{sec:res-data}

\begin{figure}[!tp]
    \centering
    \includegraphics[width=\textwidth]{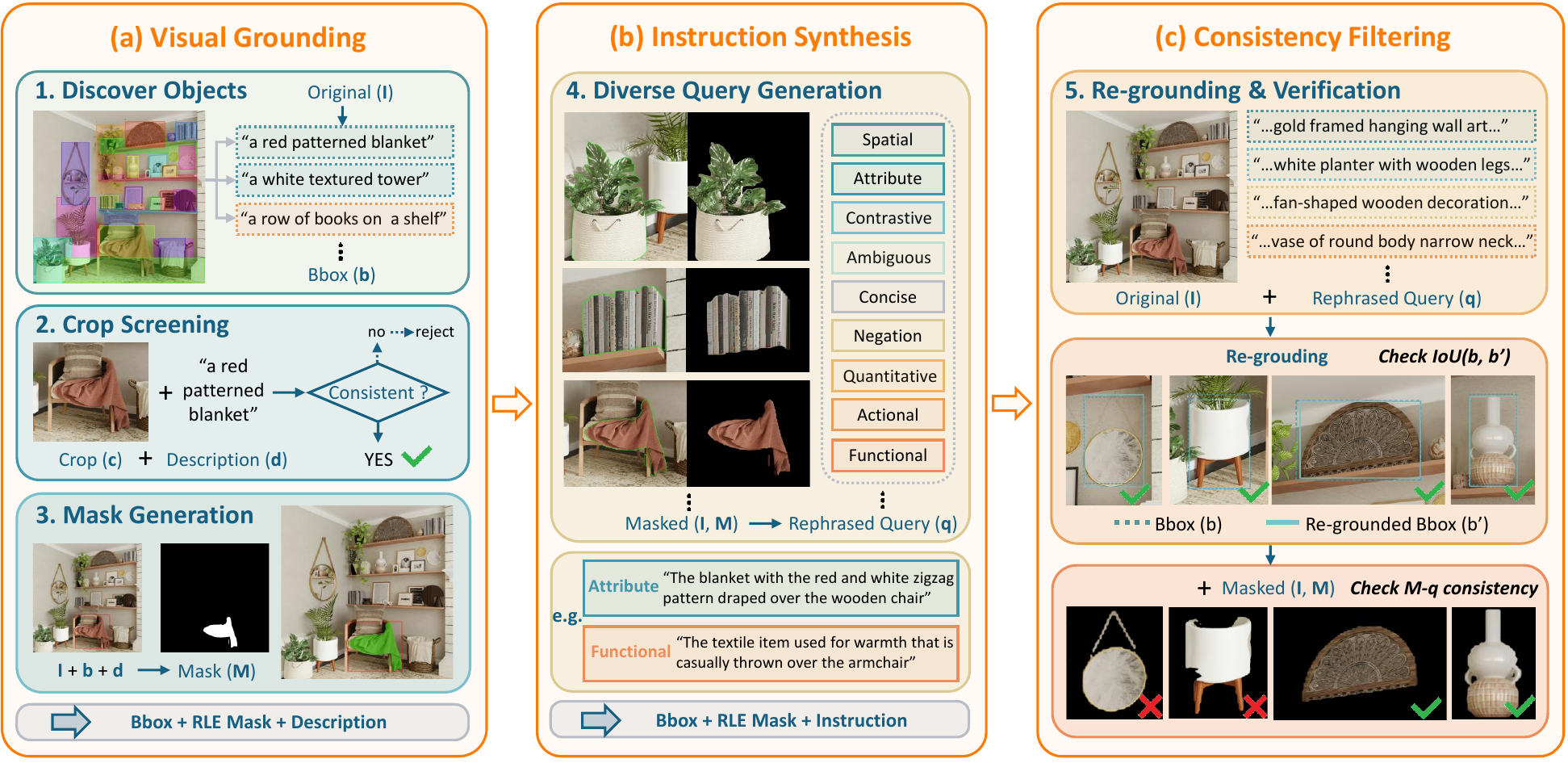}
    \caption{RES data pipeline. (a) Candidate objects are localized, their descriptions are checked against image crops, and masks are generated. (b) The masked view and original image guide referring-instruction synthesis. (c) Re-grounding and contextual verification check that each instruction identifies its intended target.}
    \label{fig:res-data-pipeline}
\end{figure}

Referring-expression data provide explicit region targets for REC and RES. We first establish a box and mask for each candidate object, then generate instructions that describe it, and finally verify that each instruction selects the annotated region (Figure \ref{fig:res-data-pipeline}). Fixing the target before instruction synthesis allows several descriptions of the same object to share a single spatial annotation.

\subsubsection{Visual Grounding}
\label{sec:res-visual-grounding}
We establish candidate regions through object discovery and crop-level verification. Given an image $I$, object discovery returns boxes $b_i$ with short descriptions $d_i$. Before generating masks or instructions, a crop--text check $\mathcal{V}_{\mathrm{crop}}(X,t)\in\{0,1\}$ verifies each description against its image crop. The accepted candidates are
\begin{equation}
    \mathcal{C}_I
    =\left\{(b_i,d_i)\in\mathcal{D}(I)
    \;\middle|\;
    \mathcal{V}_{\mathrm{crop}}\bigl(\operatorname{Crop}(I,b_i),d_i\bigr)=1
    \right\},
    \label{eq:res-visual-grounding}
\end{equation}
where $\mathcal{D}(I)$ is the set of discovered box--description pairs. For each $(b_i,d_i)\in\mathcal{C}_I$, mask generation produces $M_i=\mathcal{S}(I,b_i,d_i)\in\{0,1\}^{H\times W}$.

\subsubsection{Instruction Synthesis}
We generate referring instructions from both the target appearance and its scene context. For each retained target, a masked view $\widetilde I_i=I\odot M_i$ preserves the target pixels and sets the background to black. The generator then receives this view to identify the referent and the original image to describe its relation to surrounding objects:
\begin{equation}
    \mathcal{Q}_i=\{q_{ij}\}_{j=1}^{K_i}
    =\mathcal{G}(I,\widetilde I_i),
    \label{eq:res-instruction-synthesis}
\end{equation}
where $K_i$ is the number of expressions generated for target $i$. The expressions identify the target through location, appearance, action, function, or contrasts with neighboring objects; a blanket, for example, can be described by its pattern or its use for warmth. Concise descriptions, negation, and quantity references, such as distinguishing one object within a counted group, further diversify the wording while preserving the single annotated target.

\subsubsection{Consistency Filtering}
We retain instructions that agree with the target in both localization and meaning. Re-grounding each $q_{ij}$ in the original image must return exactly one box, $\mathcal{R}(I,q_{ij})=\{b'_{ij}\}$, whose IoU with the candidate box meets a threshold. In addition, a separate check $\mathcal{V}_{\mathrm{ref}}(I,\widetilde I_i,q_{ij})\in\{0,1\}$ uses the original image and masked view to assess whether the expression uniquely describes the target and its contextual relations. With an IoU threshold $\tau$, the retained instructions are
\begin{equation}
    \widetilde{\mathcal{Q}}_i
    =\left\{q_{ij}\in\mathcal{Q}_i
    \;\middle|\;
    \mathcal{R}(I,q_{ij})=\{b'_{ij}\},
    \;
    \operatorname{IoU}(b_i,b'_{ij})\geq\tau,
    \;\mathcal{V}_{\mathrm{ref}}(I,\widetilde I_i,q_{ij})=1
    \right\}.
    \label{eq:res-consistency-filter}
\end{equation}
The semantic check screens for references to nearby instances or properties absent from the target, complementing the box-overlap criterion. For each retained instruction, the final box is the tight bounding box of the mask, $\operatorname{box}(M_i)$. The resulting sample $(I,q_{ij},\operatorname{box}(M_i),M_i)$ pairs the same referent with box and pixel-level supervision.

\subsection{Grounded Reasoning Data}
\label{sec:reasoning-data}

\begin{figure}[!tp]
    \centering
    \includegraphics[width=\textwidth]{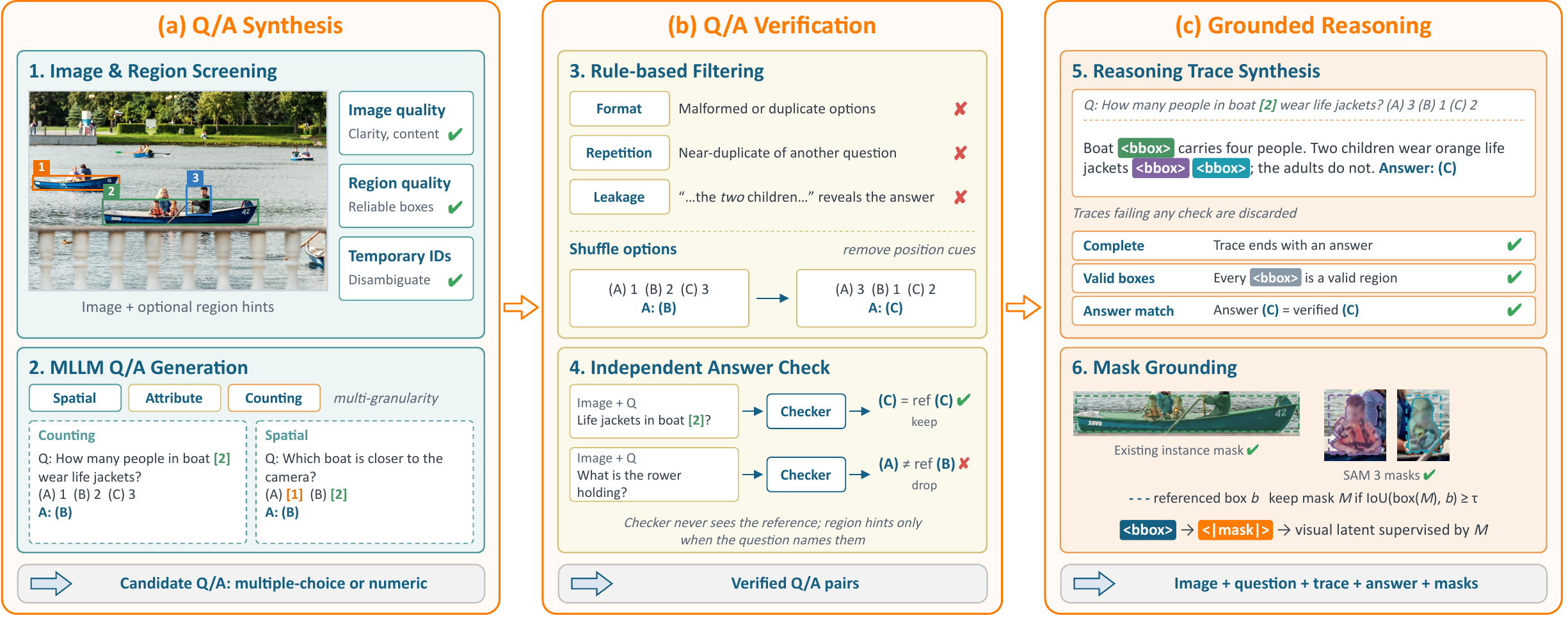}
    \caption{VQA data pipeline. (a) After image and region screening, an MLLM generates question--answer pairs with multiple-choice or numeric answers, optionally referring to regions by temporary IDs. (b) Rule-based filters remove flawed questions, options are shuffled, and an independent checker must reproduce the reference answer. (c) A reasoning model generates traces with box references. Traces that are complete, use valid boxes, and match the verified answer are paired with masks whose bounding boxes agree with the referenced boxes.}
    \label{fig:vqa-data-pipeline}
\end{figure}

Grounded reasoning data supervise the intermediate regions used in thinking with visual latents (Section \ref{sec:thinking-visual-latents}). They target the region-level failures that motivate this work, such as miscounting nearby objects and misidentifying the attributes of small regions. Each visual question answering (VQA) example pairs an image and question with a reasoning trace, a final answer, and masks for the trace's region references. We first synthesize question--answer pairs, verify the answers independently, and then construct traces with mask-annotated region references (Figure \ref{fig:vqa-data-pipeline}). The spatial training signal thus extends from final segmentation targets to the regions discussed along the reasoning trace.

\subsubsection{Question--Answer Synthesis}
We synthesize questions about counting, attributes, and spatial relations, each of which requires perceiving specific regions. After screening images and available region annotations, we assign temporary IDs to retained boxes so that questions can distinguish similar instances or combine information across regions. An MLLM generates question--answer pairs with multiple-choice or numeric answers, providing a common format for exact-match verification.

\subsubsection{Question--Answer Verification}
We verify the answers before generating reasoning traces. Rule-based filters remove malformed or duplicate options, near-duplicate questions, and wording that reveals the answer; multiple-choice options are then shuffled to remove position cues. Next, an independent checker answers each remaining question from the image without seeing the reference answer, receiving region hints only when the question names them. We retain a pair when its predicted option label or numeric value exactly matches the reference. Final examples use the original image, retaining ID markers only for regions named in the question.

\subsubsection{Trace Generation and Mask Annotation}
We generate reasoning traces that explicitly reference the image regions discussed at each intermediate step. For each verified pair, a reasoning model produces candidate traces that interleave text with \texttt{\textless{}bbox\textgreater{}} references and end with an answer. We retain complete traces with valid box references and an answer matching the verified reference. When several traces pass, we select the shortest to limit redundancy; pairs with no passing trace are discarded.

\begin{figure}[!tp]
    \centering
    \includegraphics[width=0.87\textwidth]{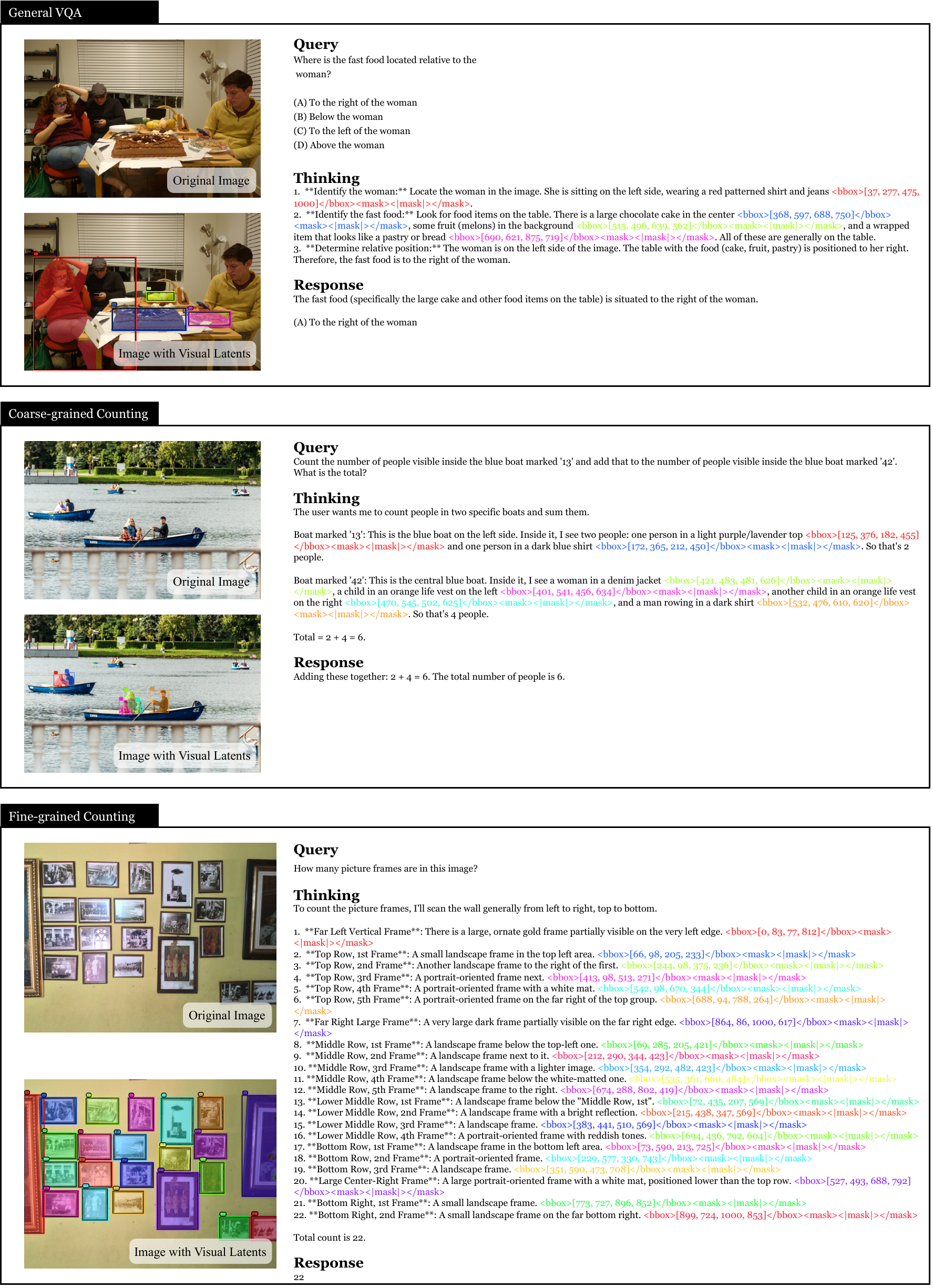}
    \caption{Thinking with visual latents. The top example identifies the position of food relative to a woman. The middle example localizes people in two boats and sums the per-boat counts. The bottom example localizes and counts individual picture frames. Colored boxes and masks show the regions referenced in each reasoning trace.}
    \label{fig:vqa-examples}
\end{figure}

After trace selection, each accepted box reference is paired with an existing instance mask or, when unavailable, a mask generated by SAM 3~\citep{carion2026sam3}. We accept the mask when the IoU between its tight bounding box and the referenced box exceeds a threshold, and append a \texttt{\textless{}|mask|\textgreater{}} token to the reference. Its contextual state becomes the visual latent supervised by that mask through Equation \ref{eq:mask-loss}. Figure \ref{fig:vqa-examples} shows counting and spatial-relation traces with masks attached to the regions named at each step.

\subsection{LensBench}
\label{sec:lensbench}

\begin{figure}[!tp]
    \centering
    \includegraphics[width=\textwidth]{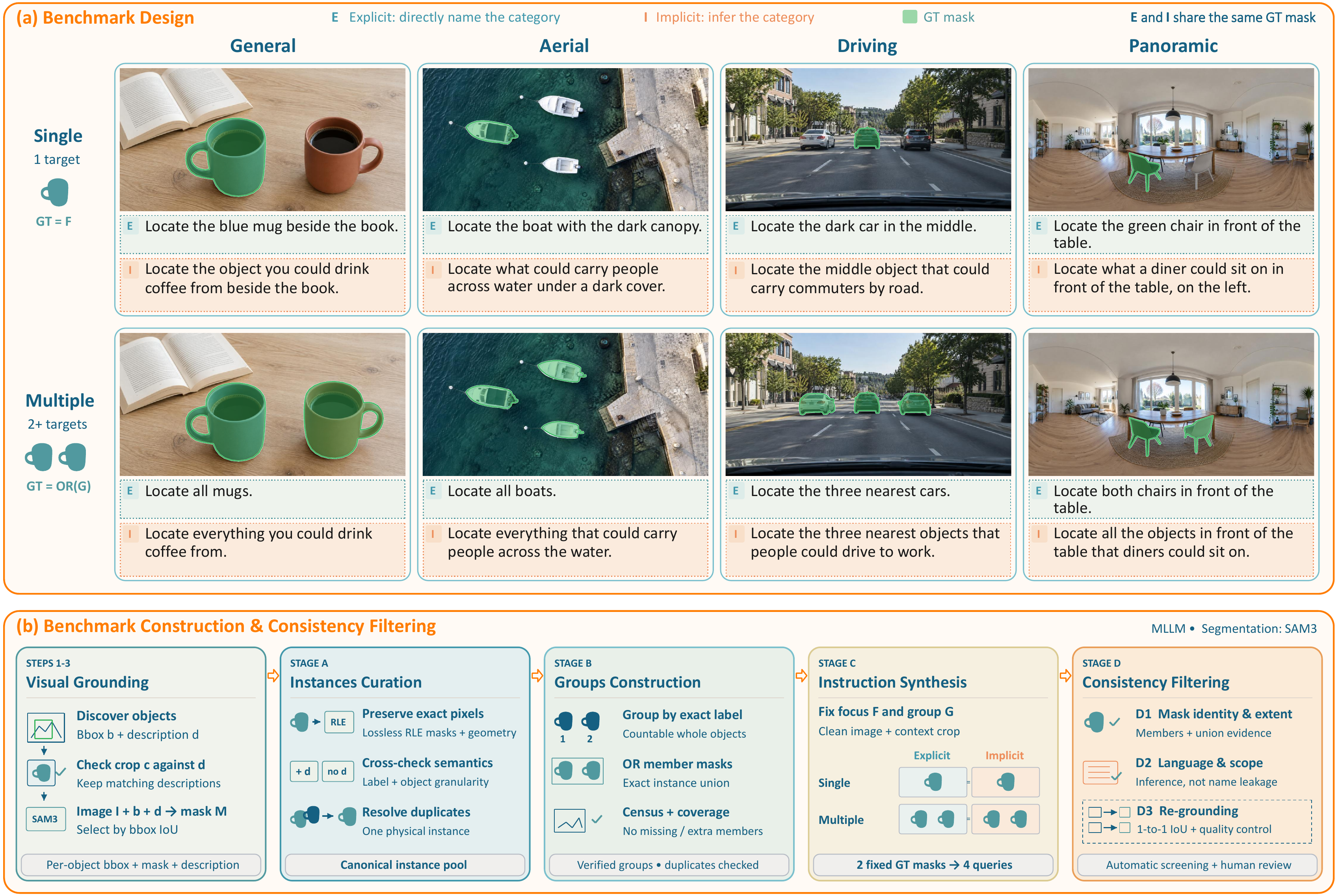}
    \caption{LensBench design and construction. (a) Examples from four domains combine single and multiple targets with explicit and implicit instructions. Each unit contains two ground-truth masks and four instructions. (b) Visual grounding, instance curation, and grouping establish the targets before instruction synthesis. Automated checks of masks, instructions, and re-grounded instances inform human review of each complete unit.}
    \label{fig:lensbench-data-pipeline}
\end{figure}

Referring benchmarks such as RefCOCO pair each expression with a single target in everyday photographs and usually name the target's category. To evaluate a broader range of referring conditions, LensBench examines how target selection and mask prediction vary when the instruction describes targets indirectly, requests several instances, or comes from a different visual domain. It covers general, aerial, driving, and panoramic imagery, with 300 complete evaluation units per domain. Each unit pairs one image with two target masks and four instructions crossing single and multiple targets with explicit and implicit wording (Figure \ref{fig:lensbench-data-pipeline}). The benchmark contains 1,200 paired units and 4,800 instruction-level samples.

\subsubsection{Benchmark Design}
Each unit separates the number of requested instances from the way they are described. Single instructions select one instance, and multiple instructions select at least two distinct instances. Explicit instructions name the category and specify the target set through attributes, relations, or quantity. Implicit instructions describe the targets through function, purpose, or commonsense relations, omitting the category and direct synonyms so that selection requires an image-supported inference.

For an image $I$, let $G$ be a set of target instances from the same category, with $|G|\geq 2$, and let $f\in G$ be a distinguishable focus instance. The target masks are
\begin{equation}
    M_{\mathrm{single}}=M_f,
    \qquad
    M_{\mathrm{multiple}}=\bigvee_{i\in G}M_i,
    \label{eq:lensbench-target-masks}
\end{equation}
where $\bigvee$ denotes pixel-wise logical OR. We generate one explicit and one implicit instruction for each mask, yielding four evaluation samples from the same image. The two instruction types share the same ground-truth mask within each cardinality, enabling paired comparisons.

\subsubsection{Benchmark Construction}
We fix the target instances and masks before generating instructions. Candidate annotations follow the visual-grounding procedure in Section \ref{sec:res-visual-grounding}; instance curation verifies identity and visible extent and resolves duplicates. Masks are stored at their original resolution using lossless run-length encoding.

Within each image, we group countable instances by category. Each group must contain every visible instance within its intended scope, without unrelated objects or duplicate physical instances. We then select a focus instance $f\in G$ and fix both target masks before generating instructions. The generator receives the original image, a crop containing the focus instance and its surroundings, and target metadata:
\begin{equation}
    \mathcal{Q}_u
    =\{q_{uj}\}_{j=1}^{4}
    =\mathcal{G}\bigl(I,\operatorname{Crop}(I,c_f),z_u\bigr),
    \label{eq:lensbench-instruction-synthesis}
\end{equation}
where $u=(I,G,f)$ indexes the unit and $c_f$ expands $\operatorname{box}(M_f)$ to include surrounding context. The indices $j=1,2,3,4$ correspond to single--explicit, single--implicit, multiple--explicit, and multiple--implicit, respectively. The metadata $z_u$ gives the canonical category, group size, and focus bounding box $\operatorname{box}(M_f)$. Each instruction must preserve its assigned target set and semantic granularity.

\subsubsection{Consistency Filtering}
Automatic screening checks mask quality, instruction semantics, and agreement with independently re-grounded instances. The mask check $\mathcal{V}_{\mathrm{mask}}(u)$ assesses identity and visible extent, detects duplicate physical instances, and verifies the union of individual masks. The instruction check $\mathcal{V}_{\mathrm{ref}}(u,q)$ uses individual and union masks to assess whether the wording is natural, self-contained, precise about the target set, and consistent with its assigned explicit or implicit condition. The intended instance set is $G_q=\{f\}$ for single and $G_q=G$ for multiple.

To check whether each instruction recovers the intended targets, re-grounding $\mathcal{R}(I,q)$ receives only the original image and instruction, with no target annotations. Its output retains duplicate predictions. We match predicted boxes to target instances one to one, maximizing total IoU, and denote the prediction assigned to instance $i$ by $b'_{qi}$. Let $\delta_q$ count pairs of identical normalized prediction boxes. With an IoU threshold $\tau$, the instructions passing all automatic checks are
\begin{equation}
    \widetilde{\mathcal{Q}}_u
    =
    \left\{
        q\in\mathcal{Q}_u
        \;\middle|\;
        \begin{aligned}
            &|\mathcal{R}(I,q)|=|G_q|,
             \quad \delta_q=0,\\
            &\min_{i\in G_q}
             \operatorname{IoU}
             \bigl(\operatorname{box}(M_i),b'_{qi}\bigr)
             \geq\tau,\\
            &\mathcal{V}_{\mathrm{mask}}(u)=1,
             \quad \mathcal{V}_{\mathrm{ref}}(u,q)=1
        \end{aligned}
    \right\}.
    \label{eq:lensbench-consistency-filter}
\end{equation}
A unit passes automatic screening when all four instructions pass, that is, $\widetilde{\mathcal{Q}}_u=\mathcal{Q}_u$. Uncertain judgments and failed requests are recorded separately to inform human review.

Finally, human reviewers make the final selection or revision decision for each complete unit. They inspect the original image, individual instance masks, both target masks, and all four instructions together to check target completeness and instruction--mask agreement. Only complete four-sample units enter LensBench, preserving the paired comparisons. Each sample is $(I,q,\operatorname{box}(M_q),M_q)$, where $M_q=\bigvee_{i\in G_q}M_i$ is the mask assigned to instruction $q$.

\FloatBarrier
\section{Experiments}
\label{sec:experiments}

The experiments test the question raised in Section \ref{sec:introduction}: whether dense mask supervision, applied to masks predicted from the MLLM's own visual representations, improves fine-grained perception and visual reasoning. We address three questions. (1) Spatial prediction: without an additional encoder for spatial features, how well does \REPORTNAME predict boxes and masks compared with methods that use one? (2) Fine-grained perception: do the gains extend to counting, attribute, and spatial-relation questions answered in text? (3) Attribution: what do mask supervision and grounded reasoning traces each contribute? Section \ref{sec:main-results} addresses the first two questions across six tasks, Section \ref{sec:lensbench-results} examines instruction semantics and visual domains on LensBench, and Section \ref{sec:ablation} addresses the third.

\subsection{Evaluation Setup}
\label{sec:evaluation}

The evaluation covers REC, RES, multi-instance grounding, reasoning segmentation, embodied referring, and fine-grained perception. All scores are percentages, and differences are reported in percentage points. Main-model results use the equal-weight average of three Stage 2 checkpoints, with a generation limit of 16,384 tokens.

\begin{itemize}
    \item \textbf{REC and RES.} We use the validation, TestA, and TestB splits of RefCOCO and RefCOCO+~\citep{yu2016modeling}, and the validation and test splits of RefCOCOg~\citep{mao2016generation,nagaraja2016modeling}. For REC, Prec@0.5 is the percentage of predicted boxes that have an IoU of at least 0.5 with the target box. A complementary box-grounding evaluation reports F1 at IoU thresholds of 0.50 and 0.95, and the mean F1 over thresholds from 0.50 to 0.95 in steps of 0.05 (F1@mIoU). RES uses cIoU on the same eight splits.

    \item \textbf{Multi-instance grounding.} COCO 2017~\citep{lin2014coco} evaluates localization of multiple object instances. We report recall, precision, and F1 at box IoU thresholds of 0.50 and 0.95, along with their mIoU aggregates.

    \item \textbf{Reasoning segmentation and embodied referring.} ReasonSeg~\citep{lai2024lisa} evaluates target selection from reasoning-dependent instructions, and RoboRefIt~\citep{lu2023vlgrasp} evaluates referring expressions in embodied scenes. We report gIoU and cIoU on the ReasonSeg validation and test sets and on RoboRefIt TestA and TestB, following the ReasonSeg convention, in which gIoU is the mean per-image IoU.

    \item \textbf{Fine-grained perception.} The English and Chinese MME-RealWorld sets (MME-RW-en and MME-RW-cn)~\citep{zhang2024mmerealworld} and CV-Bench (CV-B)~\citep{tong2024cambrian} assess general perception. V*~\citep{wu2023vstar} and the 4K and 8K HR-Bench sets (HR-4K and HR-8K)~\citep{wang2024divide} assess high-resolution perception. CountQA~\citep{tamarapalli2025countqa} and ColorBench (ColorB)~\citep{liang2025colorbench} assess counting and color recognition. All perception scores measure answer accuracy, with no box or mask output.
\end{itemize}

For $N$ image--instruction pairs with binary prediction and target masks $P_i$ and $G_i$ and nonzero unions, the benchmark definitions are
\begin{equation}
    \mathrm{gIoU}=\frac{1}{N}\sum_{i=1}^{N}
    \frac{|P_i\cap G_i|}{|P_i\cup G_i|},
    \qquad
    \mathrm{cIoU}=\frac{\sum_{i=1}^{N}|P_i\cap G_i|}
    {\sum_{i=1}^{N}|P_i\cup G_i|},
    \label{eq:segmentation-metrics}
\end{equation}
where $i$ indexes image--instruction pairs. gIoU weights each pair equally, while cIoU pools pixel intersections and unions across the dataset and therefore weights large targets more heavily.

On LensBench (Section \ref{sec:lensbench}), we report mask cIoU and Prec@0.5 by instruction type and visual domain. Unlike REC, Prec@0.5 here counts pairs whose predicted mask has an IoU of at least 0.5 with the target mask. For multiple targets, both metrics use the unions of predicted and target instance masks.

\subsection{Main Results}
\label{sec:main-results}

\subsubsection{Referring Expression Comprehension}

\REPORTNAME improves box localization over its Qwen3.5 backbone on all eight REC splits, with an average gain of 5.6 points (Table \ref{tab:referring-comprehension}). In particular, the mean gain is largest on RefCOCO+, at 7.6 points, compared with 4.4 points on both RefCOCO and RefCOCOg. RefCOCO+ excludes absolute location words~\citep{yu2016modeling}, so instances must be distinguished by appearance and attributes. Within this dataset, the TestB split shows the largest backbone-relative gain, at 9.4 points.

\begin{table}[!tp]
    \centering
    \caption{REC results on RefCOCO and RefCOCO+~\citep{yu2016modeling}, and RefCOCOg~\citep{mao2016generation,nagaraja2016modeling}. Scores are Prec@0.5 (\%, higher is better). Best scores are bold; second-best scores are underlined.}
    \resizebox{\linewidth}{!}
    {
        \begin{tabular}{lcccccccc}
        \toprule
        \multirow{2}{*}{\textbf{Methods}} & \multicolumn{3}{c}{\textbf{RefCOCO}} & \multicolumn{3}{c}{\textbf{RefCOCO+}} & \multicolumn{2}{c}{\textbf{RefCOCOg}} \\
        \cmidrule(lr){2-4}
        \cmidrule(lr){5-7}
        \cmidrule(lr){8-9}
        & Val & TestA & TestB & Val & TestA & TestB  & Val & Test \\
        \midrule
        Grounding-DINO-L~\citep{liu2024grounding} & 90.6 & 93.2 & 88.2 & 82.8 & 89.0 & 75.9 & 86.1 & 87.0 \\
        Florence-2~\citep{xiao2024florence} & 93.4 & \underline{95.3} & \underline{92.0} & 88.3 & 92.9 & 83.6 & 91.2 & 91.7 \\
        LISA$_{\text{LLaVA-7B}}$~\citep{lai2024lisa} & 85.4 & 88.8 & 82.6 & 74.2 & 79.5 & 68.4 & 79.3 & 80.4 \\
        GSVA$_{\text{Vicuna-7B}}$~\citep{xia2024gsva} & 86.3 & 89.2 & 83.8 & 72.8 & 78.8 & 68.0 & 81.6 & 81.8 \\
        PixelLM$_{\text{LLaVA-7B}}$~\citep{ren2024pixellm} & 89.8 & 92.2 & 86.4 & 83.2 & 87.0 & 78.9 & 84.6 & 86.0 \\
        VisionLLM v2$_{\text{Vicuna-7B}}$~\citep{wu2024visionllm} & 87.9 & 91.2 & 84.3 & 77.6 & 83.8 & 70.2 & 82.9 & 84.1 \\
        Text4Seg$_{\text{InternVL2-8B}}$~\citep{lan2025text4seg} & 90.3 & 93.4 & 87.5 & 85.2 & 89.9 & 79.5 & 85.4 & 85.4 \\
        UFO$_{\text{InternVL2-8B}}$~\citep{tang2025ufo} & 91.4 & 93.8 & 88.2 & 85.7 & 90.7 & 79.7 & 86.8 & 87.4 \\
        MLLMSeg$_{\text{InternVL2.5-4B}}$~\citep{wang2025mllmseg} & 91.6 & 93.3 & 87.6 & 86.0 & 90.0 & 79.8 & 88.3 & 88.6 \\
        InternVL-3.5$_{\text{4B}}$~\citep{wang2025internvl35} & 92.5 & 94.3 & 88.2 & 87.6 & 92.3 & 81.6 & 89.6 & 89.3 \\
        Qwen2.5-VL$_{\text{3B}}$~\citep{bai2025qwen2} & 89.1 & 91.7 & 84.0 & 82.4 & 88.0 & 74.1 & 85.2 & 85.7 \\
        Qwen3-VL$_{\text{4B}}$~\citep{bai2025qwen3} & 90.7 & 92.2 & 86.7 & 82.9 & 89.4 & 75.6 & 87.3 & 87.7 \\
        Qwen3-VL-Seg$_{\text{4B}}$~\citep{yao2026qwen3vlseg} & 92.7 & 94.6 & 89.8 & 87.8 & 92.1 & 82.2 & 89.1 & 89.0 \\
        Qwen3.5$_{\text{4B}}$~\citep{qwen3.5} & 90.4 & 92.3 & 87.1 & 84.8 & 88.9 & 79.1 & 88.0 & 88.2 \\
        Youtu-VL$_{\text{4B}}$~\citep{wei2026youtu} & \underline{93.6} & 95.2 & 90.8 & \underline{90.1} & \underline{93.9} & \underline{85.4} & \underline{92.2} & \textbf{92.9} \\
        \midrule
        \rowcolor{Orange}
        Xiaomi-Lens-0$_{\text{4B}}$ & \textbf{94.6} & \textbf{96.2} & \textbf{92.3} & \textbf{92.1} & \textbf{94.9} & \textbf{88.5} & \textbf{92.5} & \underline{92.4} \\
        \bottomrule
        \end{tabular}
    }
    \label{tab:referring-comprehension}
\end{table}

Among the compared models, \REPORTNAME ranks first on seven splits and second on RefCOCOg test. It exceeds Youtu-VL$_{\text{4B}}$~\citep{wei2026youtu}, the strongest baseline on each RefCOCO+ split, by 2.0, 1.0, and 3.1 points on validation, TestA, and TestB, while Youtu-VL$_{\text{4B}}$ retains a 0.5-point lead on RefCOCOg test.

\paragraph{Localization across IoU thresholds.}
The gains persist at stricter overlap thresholds (Table \ref{tab:refcoco-grounding}). F1@mIoU reaches 83.3, 81.1, and 81.1 on RefCOCO, RefCOCO+, and RefCOCOg, respectively, ranking first among the compared models and exceeding the backbone by 8.8, 11.6, and 9.5 points. Moreover, at IoU 0.95, the backbone-relative gains range from 15.5 to 16.8 points, so the improvement covers precise box placement as well as target selection.

\begin{table}[!tp]
    \centering
    \caption{REC results across IoU thresholds on RefCOCO and RefCOCO+~\citep{yu2016modeling}, and RefCOCOg~\citep{mao2016generation,nagaraja2016modeling}. F1@0.50 and F1@0.95 use the indicated IoU thresholds; F1@mIoU is the mean F1 over thresholds from 0.50 to 0.95 in steps of 0.05. All scores are percentages (higher is better). Best scores are bold; second-best scores are underlined.}
    \begingroup
    \setlength{\tabcolsep}{3pt}
    \renewcommand{\arraystretch}{1.15}
    \resizebox{\linewidth}{!}
    {
        \begin{tabular}{lccccccccc}
        \toprule
        \multirow{2}{*}{\textbf{Methods}} & \multicolumn{3}{c}{\textbf{RefCOCO}} & \multicolumn{3}{c}{\textbf{RefCOCO+}} & \multicolumn{3}{c}{\textbf{RefCOCOg}} \\
        \cmidrule(lr){2-4}\cmidrule(lr){5-7}\cmidrule(lr){8-10}
        & F1@0.50 & F1@0.95 & F1@mIoU & F1@0.50 & F1@0.95 & F1@mIoU & F1@0.50 & F1@0.95 & F1@mIoU \\
        \midrule
        GroundingDINO~\citep{liu2024grounding} & 51.2 & 23.3 & 44.3 & 49.2 & 20.9 & 41.4 & 58.0 & 23.2 & 49.7 \\
        Rex-Omni~\citep{rexomni} & 84.4 & 32.9 & 71.4 & 76.9 & 29.5 & 64.1 & 84.7 & 34.7 & 72.3 \\
        LocateAnything Fast~\citep{locateanything} & 92.5 & 44.0 & 80.6 & 84.4 & 39.5 & 73.1 & 88.5 & 40.7 & 76.2 \\
        LocateAnything Hybrid~\citep{locateanything} & \underline{92.7} & 44.3 & \underline{81.4} & 85.8 & 40.6 & \underline{75.1} & 89.4 & 41.5 & 77.7 \\
        LocateAnything Slow NTP~\citep{locateanything} & 92.4 & 38.4 & 80.2 & 85.2 & 35.7 & 73.9 & 88.7 & 35.6 & 76.1 \\
        Qwen2.5-VL$_{\text{7B}}$~\citep{bai2025qwen2} & 82.4 & 17.7 & 66.9 & 73.0 & 16.4 & 59.3 & 75.3 & 15.6 & 60.3 \\
        Qwen3-VL$_{\text{2B}}$~\citep{bai2025qwen3} & 88.3 & 26.7 & 72.4 & 81.2 & 24.3 & 66.5 & 85.9 & 25.7 & 69.8 \\
        Qwen3-VL$_{\text{4B}}$~\citep{bai2025qwen3} & 92.1 & 32.8 & 77.6 & 86.8 & 30.9 & 73.4 & 88.8 & 32.3 & 74.7 \\
        Qwen3-VL$_{\text{8B}}$~\citep{bai2025qwen3} & 91.1 & 29.9 & 75.8 & 86.4 & 29.0 & 72.0 & 88.7 & 28.8 & 73.3 \\
        Qwen3.5$_{\text{4B}}$~\citep{qwen3.5} & 90.1 & 29.1 & 74.5 & 84.1 & 27.8 & 69.5 & 88.4 & 27.0 & 71.6 \\
        Qwen3.5$_{\text{9B}}$~\citep{qwen3.5} & 92.0 & 35.1 & 78.3 & \underline{87.5} & 33.9 & 74.6 & \underline{89.5} & 33.5 & 75.3 \\
        RynnBrain1.1~\citep{li2026rynnbrain11} & 82.6 & 20.9 & 65.8 & 72.5 & 19.2 & 57.5 & 84.3 & 21.5 & 67.6 \\
        SenseNova-Vision~\citep{han2026sensenovavision} & 90.1 & \textbf{45.1} & 79.7 & 84.4 & \underline{41.7} & 74.3 & 89.2 & \textbf{44.7} & \underline{78.7} \\
        MiMo-VL-SFT$_{\text{7B}}$~\citep{xiaomi2025mimovl} & 86.1 & 14.5 & 66.3 & 76.8 & 13.1 & 58.9 & 84.9 & 14.3 & 64.8 \\
        MiMo-VL-RL$_{\text{7B}}$~\citep{xiaomi2025mimovl} & 90.4 & 14.2 & 68.5 & 84.8 & 13.6 & 64.4 & 87.7 & 13.1 & 64.8 \\
        BAGEL~\citep{deng2025bagel} & 79.4 & 25.1 & 65.2 & 68.3 & 20.5 & 54.5 & 77.7 & 22.2 & 62.3 \\
        RynnBrain~\citep{dang2026rynnbrain} & 74.0 & 13.6 & 55.6 & 65.5 & 11.7 & 48.9 & 76.5 & 14.3 & 57.9 \\
        \midrule
        \rowcolor{Orange}
        Xiaomi-Lens-0$_{\text{4B}}$  & \textbf{94.6} & \underline{44.6} & \textbf{83.3} & \textbf{92.1} & \textbf{43.6} & \textbf{81.1} & \textbf{92.5} & \underline{43.8} & \textbf{81.1} \\
        \bottomrule
        \end{tabular}
    }
    \endgroup
    \label{tab:refcoco-grounding}
\end{table}

\subsubsection{Referring Expression Segmentation}

\begin{table}[!tp]
    \centering
    \caption{RES results on RefCOCO and RefCOCO+~\citep{yu2016modeling}, and RefCOCOg~\citep{mao2016generation,nagaraja2016modeling}. Scores are cIoU (\%, higher is better). Best scores are bold; second-best scores are underlined.}
    \resizebox{\linewidth}{!}
    {
         \begin{tabular}{lcccccccc}
        \toprule
        \multirow{2}{*}{\textbf{Methods}} & \multicolumn{3}{c}{\textbf{RefCOCO}} & \multicolumn{3}{c}{\textbf{RefCOCO+}} & \multicolumn{2}{c}{\textbf{RefCOCOg}} \\
        \cmidrule(lr){2-4}
        \cmidrule(lr){5-7}
        \cmidrule(lr){8-9}
        & Val & TestA & TestB & Val & TestA & TestB  & Val & Test \\
        \midrule
        LISA$_{\text{LLaVA-7B}}$~\citep{lai2024lisa} & 74.9 & 79.1 & 72.3 & 65.1 & 70.8 & 58.1 & 67.9 & 70.6 \\
        GSVA$_{\text{Vicuna-7B}}$~\citep{xia2024gsva} & 77.2 & 78.9 & 73.5 & 65.9 & 69.6 & 59.8 & 72.7 & 73.3 \\
        PixelLM$_{\text{LLaVA-7B}}$~\citep{ren2024pixellm} & 73.0 & 76.5 & 68.2 & 66.3 & 71.7 & 58.3 & 69.3 & 70.5 \\
        VisionLLM v2$_{\text{Vicuna-7B}}$~\citep{wu2024visionllm} & 76.6 & 79.3 & 74.3 & 64.5 & 69.8 & 61.5 & 70.7 & 71.2 \\
        Text4Seg$_{\text{InternVL2-8B}}$~\citep{lan2025text4seg} & 79.2 & 81.7 & 75.6 & 72.8 & 77.9 & 66.5 & 74.0 & 75.3 \\
        UFO$_{\text{InternVL2-8B}}$~\citep{tang2025ufo} & 78.0 & 79.7 & 75.6 & 72.3 & 76.8 & 66.6 & 73.7 & 74.3 \\
        MLLMSeg$_{\text{InternVL2.5-4B}}$~\citep{wang2025mllmseg} & 79.5 & 81.4 & 77.7 & 75.4 & 78.7 & 70.8 & 78.1 & \underline{78.4} \\
        SAM 3~\citep{carion2026sam3} & 75.5 & 77.6 & 71.0 & 67.3 & 71.1 & 63.4 & 73.4 & 74.0 \\
        UniPixel$_{\text{Qwen2.5-VL-3B}}$~\citep{liu2025unipixel} & 80.5 & 82.6 & 76.9 & 74.3 & 78.9 & 68.4 & 76.3 & 77.0 \\
        Youtu-VL$_{\text{4B}}$~\citep{wei2026youtu} & 80.7 & 82.0 & 78.4 & \underline{76.2} & 79.6 & \underline{71.4} & 76.5 & 76.6 \\
        Qwen3-VL-Seg$_{\text{4B}}$~\citep{yao2026qwen3vlseg} & \underline{82.3} & \underline{83.7} & \underline{79.1} & \underline{76.2} & \underline{80.2} & 70.8 & \underline{78.2} & 78.1 \\
        \midrule
        \rowcolor{Orange}
        Xiaomi-Lens-0$_{\text{4B}}$ & \textbf{83.4} & \textbf{84.6} & \textbf{81.2} & \textbf{79.9} & \textbf{83.3} & \textbf{76.3} & \textbf{81.1} & \textbf{81.2} \\
        \bottomrule
        \end{tabular}

    }
    \label{tab:referring-segmentation}
\end{table}

RES directly tests whether masks decoded from the shared patch-level grid can match methods that obtain spatial features from an additional encoder. Under this setup, \REPORTNAME achieves the highest cIoU among the compared models on all eight RES splits (Table \ref{tab:referring-segmentation}), averaging 81.4 versus 78.6 for Qwen3-VL-Seg$_{\text{4B}}$~\citep{yao2026qwen3vlseg}, the strongest baseline on average. Margins over the best competing score on each split range from 0.9 to 2.1 points on RefCOCO, 3.1 to 4.9 on RefCOCO+, and 2.8 to 2.9 on RefCOCOg. \REPORTNAME also exceeds the dedicated segmentation model SAM 3~\citep{carion2026sam3} by 7.0 to 12.9 points.

MLLMSeg$_{\text{InternVL2.5-4B}}$~\citep{wang2025mllmseg} also decodes masks from the MLLM's vision-encoder features, and \REPORTNAME exceeds it by 2.8 to 5.5 points across the eight splits. On both RefCOCOg RES splits, \REPORTNAME also exceeds Youtu-VL$_{\text{4B}}$ by 4.6 cIoU points. Figures \ref{fig:rec-cases} and \ref{fig:res-cases} in Appendix \ref{sec:case-study} illustrate the resulting box and mask predictions.

\subsubsection{Multi-Instance Grounding}
\label{sec:multi-instance-results}

\begin{table}[!tp]
    \centering
    \caption{Multi-instance grounding results on COCO 2017~\citep{lin2014coco}. Recall, precision, and F1 are reported at IoU thresholds of 0.50 and 0.95 and as mIoU aggregates. All scores are percentages. Best scores are bold; second-best scores are underlined.}
    \begingroup
    \setlength{\tabcolsep}{3pt}
    \renewcommand{\arraystretch}{1.10}
    \resizebox{\linewidth}{!}
    {
        \begin{tabular}{lccccccccc}
        \toprule
        \multirow{2}{*}{\textbf{Methods}} & \multicolumn{3}{c}{\textbf{IoU 0.50}} & \multicolumn{3}{c}{\textbf{IoU 0.95}} & \multicolumn{3}{c}{\textbf{mIoU}} \\
        \cmidrule(lr){2-4}\cmidrule(lr){5-7}\cmidrule(lr){8-10}
        & Recall & Precision & F1 & Recall & Precision & F1 & Recall & Precision & F1 \\
        \midrule
        GroundingDINO~\citep{liu2024grounding} & \underline{79.6} & \textbf{83.2} & \textbf{81.4} & 24.7 & 28.1 & 26.3 & 61.1 & 63.7 & 60.6 \\
        Rex-Omni~\citep{rexomni} & 69.9 & 81.0 & 75.0 & 17.9 & 21.7 & 19.7 & 53.9 & 59.4 & 56.3 \\
        LocateAnything Fast~\citep{locateanything} & 66.2 & 73.3 & 69.6 & 24.7 & 26.9 & 25.8 & 50.9 & 59.3 & 53.1 \\
        LocateAnything Hybrid~\citep{locateanything} & 74.7 & 80.8 & 77.6 & \underline{26.6} & \underline{28.7} & \underline{27.6} & 52.4 & \textbf{65.2} & 59.1 \\
        LocateAnything Slow NTP~\citep{locateanything} & 75.0 & \underline{81.9} & 78.3 & 25.0 & 27.2 & 26.0 & 54.7 & \underline{65.1} & 59.4 \\
        Qwen2.5-VL$_{\text{7B}}$~\citep{bai2025qwen2} & 57.2 & 62.9 & 59.9 & 12.2 & 13.2 & 12.7 & 44.7 & 49.1 & 46.3 \\
        Qwen3-VL$_{\text{2B}}$~\citep{bai2025qwen3} & 63.5 & 67.1 & 65.3 & 21.8 & 22.6 & 22.2 & 42.6 & 44.9 & 42.5 \\
        Qwen3-VL$_{\text{4B}}$~\citep{bai2025qwen3} & 66.1 & 70.6 & 68.3 & 23.7 & 24.9 & 24.3 & 44.9 & 47.8 & 46.5 \\
        Qwen3-VL$_{\text{8B}}$~\citep{bai2025qwen3} & 65.7 & 69.5 & 67.5 & 23.9 & 25.1 & 24.5 & 44.8 & 47.3 & 46.3 \\
        RynnBrain1.1~\citep{li2026rynnbrain11} & 36.5 & 60.5 & 45.5 & 11.6 & 16.5 & 13.6 & 28.7 & 45.3 & 35.2 \\
        SenseNova-Vision~\citep{han2026sensenovavision} & 79.1 & 80.5 & 79.8 & \textbf{27.2} & \textbf{30.2} & \textbf{28.6} & 63.1 & 55.4 & 57.5 \\
        GroundingPI~\citep{yu2026groundingpi} & \textbf{81.4} & 80.9 & \underline{81.2} & 22.9 & 22.8 & 22.8 & \underline{63.2} & 62.8 & \underline{63.0} \\
        MiMo-VL-SFT$_{\text{7B}}$~\citep{xiaomi2025mimovl} & 65.5 & 72.2 & 68.7 & 10.0 & 11.1 & 10.5 & 45.7 & 50.5 & 48.0 \\
        MiMo-VL-RL$_{\text{7B}}$~\citep{xiaomi2025mimovl} & 65.6 & 74.5 & 69.8 & 9.2 & 10.3 & 9.7 & 44.5 & 50.2 & 47.2 \\
        BAGEL~\citep{deng2025bagel} & 65.6 & 75.2 & 70.1 & 11.0 & 13.0 & 11.9 & 46.3 & 47.2 & 46.0 \\
        RynnBrain~\citep{dang2026rynnbrain} & 21.3 & 49.9 & 29.8 & 4.9 & 10.9 & 6.7 & 15.5 & 35.4 & 21.5 \\
        \midrule
        \rowcolor{Orange}
        Xiaomi-Lens-0$_{\text{4B}}$ & \underline{79.6} & 81.1 & 80.4 & \textbf{27.2} & 27.7 & 27.4 & \textbf{63.4} & 64.5 & \textbf{63.9} \\
        \bottomrule
        \end{tabular}
    }
    \endgroup
    \label{tab:coco-grounding}
\end{table}

Beyond single-target referring, \REPORTNAME also localizes multiple object instances on COCO 2017 (Table \ref{tab:coco-grounding}). Under the mIoU aggregate, it achieves the highest F1 among the compared models, at 63.9, which is 0.9 points above the grounding foundation model GroundingPI~\citep{yu2026groundingpi} and 3.3 points above GroundingDINO~\citep{liu2024grounding}. At IoU 0.50, its F1 of 80.4 trails only these two models. However, the order reverses at IoU 0.95: its F1 of 27.4 exceeds GroundingPI by 4.6 points and GroundingDINO by 1.1 points, trailing only SenseNova-Vision~\citep{han2026sensenovavision} and LocateAnything Hybrid~\citep{locateanything}. Therefore, its lead over GroundingPI comes from tighter box placement at strict thresholds rather than from recovering more instances. Figure \ref{fig:coco-cases} illustrates predictions for repeated and overlapping objects.

\subsubsection{Reasoning Segmentation and Embodied Referring}

\begin{table}[!tp]
    \centering
    \caption{Reasoning segmentation on ReasonSeg~\citep{lai2024lisa} and embodied referring on RoboRefIt~\citep{lu2023vlgrasp}. Scores are gIoU and cIoU (\%, higher is better; Equation \ref{eq:segmentation-metrics}). Best scores are bold; second-best scores are underlined.}
    \resizebox{\linewidth}{!}
    {
        \begin{tabular}{lcccccccc}
        \toprule
        \multirow{3}{*}{\textbf{Methods}} & \multicolumn{4}{c}{\textbf{ReasonSeg}} & \multicolumn{4}{c}{\textbf{RoboRefIt}} \\
        \cmidrule(lr){2-5}
        \cmidrule(lr){6-9}
        & \multicolumn{2}{c}{Val} & \multicolumn{2}{c}{Test} & \multicolumn{2}{c}{TestA} & \multicolumn{2}{c}{TestB} \\
        \cmidrule(lr){2-3}
        \cmidrule(lr){4-5}
        \cmidrule(lr){6-7}
        \cmidrule(lr){8-9}
        & gIoU & cIoU & gIoU & cIoU & gIoU & cIoU & gIoU & cIoU \\
        \midrule
        LISA$_{\text{LLaVA-7B}}$~\citep{lai2024lisa} & 52.9 & 54.0 & 47.3 & 48.4 & 36.1 & 29.9 & 28.7 & 22.7\\
        LISA++$_{\text{LLaVA-7B}}$~\citep{yang2023lisa++} & 64.2 & \underline{68.1} & 57.0 & 59.5 & 58.7 & 51.5 & 53.5 & 47.8 \\
        PixelLM$_{\text{LLaVA-7B}}$~\citep{ren2024pixellm} & 44.7 & 37.4 & 44.0 & 39.7 & 45.2 & 38.4 & 38.2 & 31.4\\
        Sa2VA$_{\text{InternVL2-2B}}$~\citep{yuan2025sa2va} & 59.2 & 60.0 & 56.9 & 55.8 & 56.8 & 29.0 & 34.0 & 10.2\\
        Sa2VA$_{\text{InternVL2-8B}}$~\citep{yuan2025sa2va} & \underline{65.2} & 61.1 & 59.5 & 57.8 & 58.7 & 27.9 & 36.1 & 9.0\\
        X-SAM$_{\text{Phi-3-3.8B}}$~\citep{wang2026x} & 56.6 & 32.9 & 57.8 & 41.0 & 41.2 & 10.3 & 25.1 & 4.0\\
        InstructSAM$_{\text{Qwen3-VL-2B}}$~\citep{yuan2026instructsam} & 62.5 & 65.0 & \underline{61.1} & \underline{61.0} & \underline{82.5} & \underline{58.4} & \underline{74.4} & \underline{55.5}\\
        \midrule
        \rowcolor{Orange}
        Xiaomi-Lens-0$_{\text{4B}}$ & \textbf{71.2} & \textbf{68.5} & \textbf{68.0} & \textbf{62.5} & \textbf{87.0} & \textbf{88.4} & \textbf{81.5} & \textbf{81.2} \\
        
        \bottomrule
        \end{tabular}
    }
    \label{tab:reasoning-embodied}
\end{table}

Target selection becomes harder when the instruction identifies the target through its function or scene context. \REPORTNAME leads the compared models in both gIoU and cIoU on the two ReasonSeg splits (Table \ref{tab:reasoning-embodied}). On the test split, it reaches 68.0 gIoU and 62.5 cIoU, exceeding the strongest baseline, Qwen3-VL-based InstructSAM~\citep{yuan2026instructsam}, by 6.9 and 1.5 points. On validation, the cIoU margin is 0.4 points over LISA++$_{\text{LLaVA-7B}}$~\citep{yang2023lisa++}, at 68.5 versus 68.1.

For embodied referring on RoboRefIt, \REPORTNAME reaches 87.0 gIoU and 88.4 cIoU on TestA, and 81.5 gIoU and 81.2 cIoU on TestB. It exceeds InstructSAM, the strongest listed gIoU baseline, by 4.5 and 7.1 points on the respective splits. Figures \ref{fig:reasonseg-cases} and \ref{fig:roborefit-cases} illustrate target selection through function, appearance, and spatial context.

\subsubsection{Fine-Grained Perception}
\begin{table}[!tp]
    \centering
    \caption{General and fine-grained perception results. All scores are answer accuracy (\%, higher is better). Benchmark names and abbreviations are defined in the evaluation setup. Best scores are bold; second-best scores are underlined.}
    \resizebox{\linewidth}{!}
    {
        \begin{tabular}{lcccccccc}
        \toprule
        \multirow{2}{*}{\textbf{Methods}} & \multicolumn{3}{c}{\textbf{General Perception}} & \multicolumn{3}{c}{\textbf{High-Resolution Perception}} & \multicolumn{2}{c}{\textbf{Counting and Color}} \\
        \cmidrule(lr){2-4} 
        \cmidrule(lr){5-7} 
        \cmidrule(lr){8-9}
        & MME-RW-en & MME-RW-cn & CV-B & V* & HR-4K & HR-8K & CountQA & ColorB \\
        \midrule
        Qwen2.5-VL$_{\text{7B}}$~\citep{bai2025qwen2} & 60.8 & 58.3 & 75.3 & 78.5 & 71.6 & 67.9 & 18.9 & 76.4 \\
        MiMo-VL-RL$_{\text{7B}}$~\citep{xiaomi2025mimovl} & 63.4 & 59.8 & 84.3 & 81.2 & 74.4 & 72.9 & 28.3 & \underline{82.8} \\
        Qwen3-VL$_{\text{4B}}$~\citep{bai2025qwen3} & 63.5 & 63.6 & 85.0 & 80.1 & 78.3 & 72.9 & 28.1 & 81.6 \\
        Qwen3-VL$_{\text{8B}}$~\citep{bai2025qwen3} & 66.0 & \underline{66.7} & \underline{85.4} & \underline{86.4} & 78.9 & 74.6 & 29.0 & \underline{82.8} \\
        MiniCPM-V-4.5$_{\text{9B}}$~\citep{yu2025minicpmv45} & 58.2 & 56.2 & 80.2 & 70.2 & 69.9 & 63.6 & 23.4 & 79.8 \\
        Qwen3.5$_{\text{4B}}$~\citep{qwen3.5} & \underline{66.5} & 60.3 & \underline{85.4} & 82.2 & \underline{81.5} & \underline{75.6} & \underline{43.8} & 82.7 \\
        \midrule
        \rowcolor{Orange}
        Xiaomi-Lens-0$_{\text{4B}}$ & \textbf{69.7} & \textbf{67.5} & \textbf{88.1} & \textbf{87.4} & \textbf{82.5} & \textbf{80.6} & \textbf{50.6} & \textbf{84.4} \\
        \bottomrule
        \end{tabular}
    }
    \label{tab:perception}
\end{table}

To test whether the gains extend beyond spatial outputs, the perception benchmarks evaluate the region-level failures identified in Section \ref{sec:introduction} through text answers alone. \REPORTNAME exceeds its Qwen3.5$_{\text{4B}}$ backbone and every other compared open-source model on all eight benchmarks in Table \ref{tab:perception}. Section \ref{sec:ablation} isolates the contribution of mask supervision to these gains.

\paragraph{Counting and color recognition.}
CountQA and ColorB directly test counting and attribute recognition. CountQA accuracy rises from 43.8 to 50.6, a 6.8-point gain over the backbone, and ColorB improves by 1.7 points to 84.4. Figures \ref{fig:hrbench-cases} and \ref{fig:countqa-cases} show thinking with visual latents on such questions, where region references in the reasoning trace link intermediate statements to decoded masks (Appendix \ref{sec:case-study}).

\paragraph{General and high-resolution perception.}
Backbone-relative gains are 3.2 points on MME-RW-en, 7.2 on MME-RW-cn, and 2.7 on CV-B. On the high-resolution benchmarks, which require recognizing small regions in large images, gains range from 1.0 point on HR-4K to 5.2 on V*, with a 5.0-point improvement on HR-8K. \REPORTNAME also exceeds Qwen3-VL$_{\text{8B}}$~\citep{bai2025qwen3} on all three high-resolution sets.

\subsection{LensBench Results}
\label{sec:lensbench-results}

\begin{table}[!tp]
    \centering
    \caption{LensBench marginal results by instruction type and visual domain, pooling single and multiple targets. Scores are mask cIoU and Prec@0.5 (\%, higher is better); Prec@0.5 uses a mask IoU threshold of 0.5, and multiple targets use the unions of predicted and target instance masks. All models are evaluated by us with the same instructions and predict masks directly. Best scores are bold; second-best scores are underlined.}
    \begingroup
    \setlength{\tabcolsep}{3pt}
    \renewcommand{\arraystretch}{1.10}
    \resizebox{\linewidth}{!}
    {
        \begin{tabular}{lcccccccccccc}
        \toprule
        \multirow{3}{*}{\textbf{Methods}} & \multicolumn{4}{c}{\textbf{Instruction}}& \multicolumn{8}{c}{\textbf{Domain}} \\
        \cmidrule(lr){2-5}
        \cmidrule(lr){6-13}
        & \multicolumn{2}{c}{Explicit} & \multicolumn{2}{c}{Implicit} & \multicolumn{2}{c}{General} & \multicolumn{2}{c}{Aerial} & \multicolumn{2}{c}{Driving} & \multicolumn{2}{c}{Panoramic} \\
        \cmidrule(lr){2-3}
        \cmidrule(lr){4-5}
        \cmidrule(lr){6-7}
        \cmidrule(lr){8-9}
        \cmidrule(lr){10-11}
        \cmidrule(lr){12-13}
        & cIoU & Prec@0.5 & cIoU & Prec@0.5 & cIoU & Prec@0.5 & cIoU & Prec@0.5 & cIoU & Prec@0.5 & cIoU & Prec@0.5 \\
        \midrule
        GPT-5.6-Sol~\citep{openai2026gpt56sol} & 63.8 & 73.2 & \underline{61.2} & \underline{70.0} & \underline{68.4} & 67.9 & 42.2 & 58.5 & \underline{68.4} & \textbf{81.8} & 57.1 & \underline{78.3} \\
        Gemini-3.6-Flash~\citep{google2026gemini36flash} & 61.4 & \underline{74.5} & 61.0 & \textbf{70.3} & 63.6 & 65.8 & \underline{50.1} & \underline{62.5} & 66.8 & \underline{81.0} & \underline{60.8} & \textbf{80.3} \\
        MiMo-V2.5~\citep{xiaomi2026mimov25} & 51.4 & 52.2 & 47.4 & 46.8 & 54.2 & 45.2 & 32.6 & 38.6 & 50.5 & 52.8 & 45.4 & 61.4 \\
        Seed-2.1-Pro~\citep{bytedance2026seed21} & 62.1 & 70.2 & 55.9 & 64.5 & 60.6 & 56.7 & \underline{50.1} & 58.5 & 66.8 & 77.4 & 59.9 & 76.7 \\
        Qwen3.8-Max~\citep{qwen2026qwen38} & \underline{67.6} & 69.2 & \textbf{64.5} & 63.8 & \textbf{74.2} & \textbf{79.7} & 33.5 & \underline{62.5} & 58.0 & 57.7 & \underline{60.8} & 66.2 \\
        Kimi-K3~\citep{kimi2026k3} & 29.4 & 39.1 & 28.3 & 33.0 & 38.5 & 26.1 & 13.8 & 44.9 & 47.2 & 31.1 & 13.4 & 42.1 \\
        GLM-5.3-Flash~\citep{zai2026glm53flash} & 44.3 & 37.5 & 43.5 & 33.3 & 48.9 & 26.3 & 28.0 & 33.4 & 51.6 & 47.5 & 30.6 & 34.4 \\
        DeepSeek-V4-Flash-Vision-Exp~\citep{deepseek2026v4flashvision} & 41.8 & 30.7 & 41.1 & 27.3 & 46.1 & 28.1 & 25.9 & 28.9 & 43.9 & 27.9 & 32.8 & 31.0 \\
        SAM 3~\citep{carion2026sam3} & 17.3 & 38.1 & 15.4 & 19.5 & 23.8 & 24.8 & 4.3 & 27.3 & 27.7 & 35.1 & 11.1 & 27.9 \\
        \midrule
        \rowcolor{Orange}
        Xiaomi-Lens-0$_{\text{4B}}$ & \textbf{71.4} & \textbf{85.0} & 54.7 & \underline{70.0} & 63.9 & \underline{79.1} & \textbf{57.1} & \textbf{79.6} & \textbf{68.6} & 79.4 & \textbf{62.3} & 72.1 \\
        \bottomrule
        \end{tabular}
    }
    \endgroup
    \label{tab:lensbench}
\end{table}

LensBench examines region prediction as instruction semantics and visual domains vary. Table \ref{tab:lensbench} reports mask cIoU and Prec@0.5 by instruction type and domain, pooling single and multiple targets. The evaluation covers 300 samples in each of the 16 domain--cardinality--instruction partitions, totaling 4,800 instructions from 1,200 paired units. Explicit and implicit wording share the same target mask within each cardinality. We evaluate all compared models under the same instructions, and each model predicts masks directly from the image and instruction.

\REPORTNAME achieves the highest cIoU among the compared models in three of the four domains: 57.1 on aerial, 68.6 on driving, and 62.3 on panoramic. The corresponding margins over the strongest baseline in each domain are 7.0, 0.2, and 1.5 points. Aerial imagery shows the largest gains in both metrics: the strongest baselines reach 50.1 cIoU and 62.5 Prec@0.5, compared with 57.1 and 79.6 for \REPORTNAME. By contrast, on general imagery, Qwen3.8-Max~\citep{qwen2026qwen38} leads with 74.2 cIoU and 79.7 Prec@0.5, while \REPORTNAME reaches 63.9 and 79.1.

Across instruction types, implicit instructions remain harder than explicit ones. \REPORTNAME leads on explicit instructions with 71.4 cIoU and 85.0 Prec@0.5. However, on implicit instructions, its cIoU falls to 54.7, ranking fifth behind Qwen3.8-Max at 64.5, GPT-5.6-Sol~\citep{openai2026gpt56sol} at 61.2, Gemini-3.6-Flash~\citep{google2026gemini36flash} at 61.0, and Seed-2.1-Pro~\citep{bytedance2026seed21} at 55.9. By contrast, its Prec@0.5 of 70.0 ties GPT-5.6-Sol for second behind Gemini-3.6-Flash at 70.3. The decline from explicit to implicit instructions is 16.7 points in cIoU and 15.0 points in Prec@0.5.

Examining instruction type and cardinality together reveals a particularly large drop on panoramic imagery with multiple targets. Prec@0.5 falls from 81.3 for explicit instructions to 48.7 for implicit instructions, and cIoU falls from 71.8 to 48.9. These results identify implicit selection of multiple targets in panoramic scenes as a remaining challenge.

\begin{figure}[!tp]
    \centering
    \includegraphics[width=\textwidth]{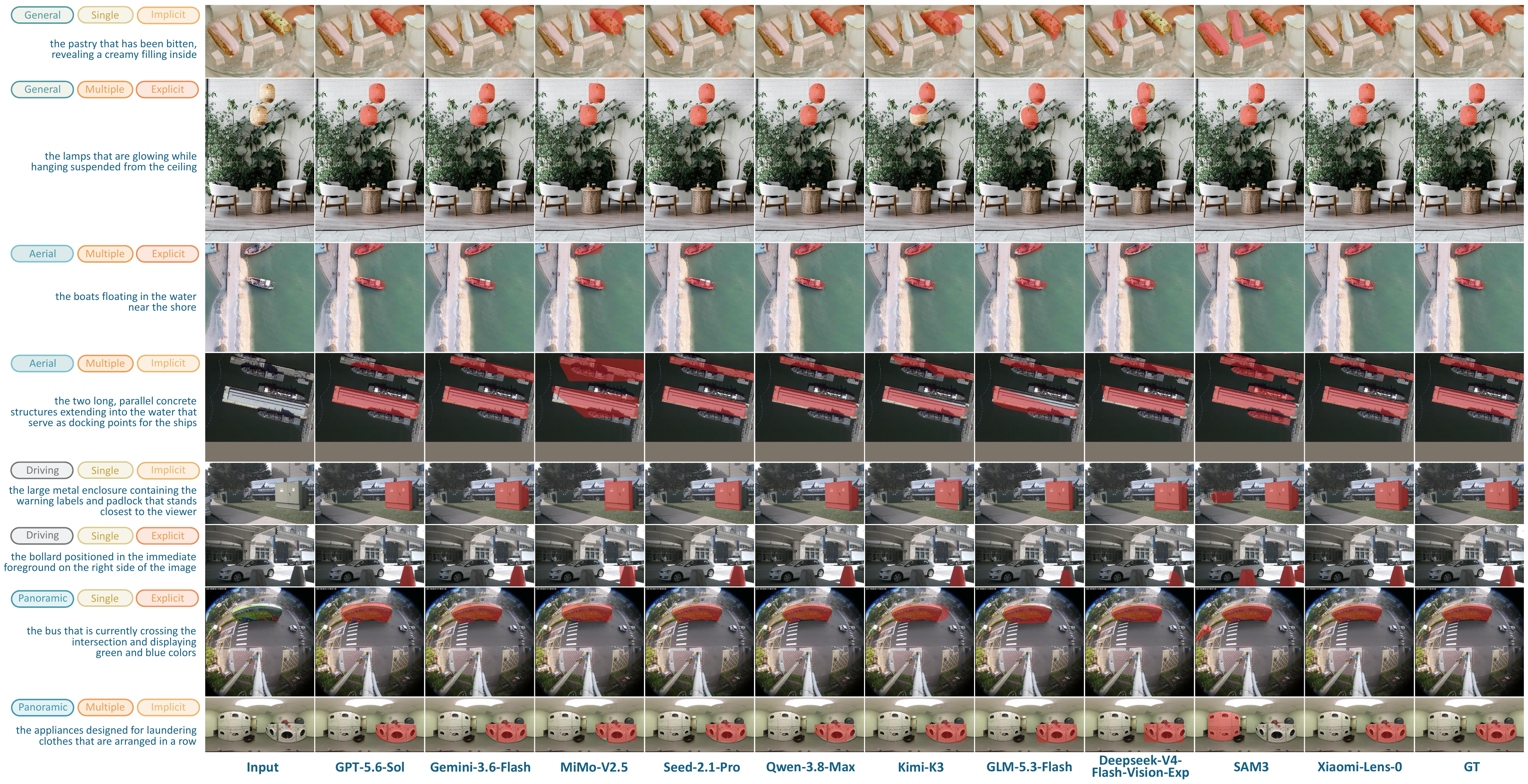}
    \caption{Qualitative comparison on LensBench across general, aerial, driving, and panoramic imagery. Each row gives the domain, target cardinality, and instruction type alongside the query. Columns show the input image, predictions from the compared models, and the ground truth (GT); red overlays indicate the selected regions. The examples illustrate attribute- and function-based target selection, recovery of multiple instances, and differences in mask extent.}
    \label{fig:lensbench-evaluation}
\end{figure}

\paragraph{Instruction-conditioned target selection.}
Figure \ref{fig:lensbench-evaluation} complements the aggregate results with visual comparisons under the same instructions. In general imagery, \REPORTNAME selects the bitten pastry using its exposed filling, while SAM 3 selects several other pastries. In driving scenes, it identifies the foreground metal enclosure from its warning labels, padlock, and relative position, and distinguishes the requested right-hand bollard from nearby posts. Moreover, the panoramic bus example shows a mask that follows the target's curved appearance under image distortion. These cases illustrate how appearance, spatial context, and instruction semantics jointly determine the selected region across domains.

\paragraph{Multiple targets and region extent.}
The multi-target examples distinguish recovering the requested set from delineating its members. \REPORTNAME covers both hanging lamps and both boats near the shore. For the implicit aerial query, it selects the two parallel docking structures while separating them from the ships and intervening water. In comparison, SAM 3 also includes ships, and MiMo-V2.5 extends the selected region into the water between the structures. This example illustrates the mask-level differences behind the aerial results: identifying the vicinity of the targets still leaves their spatial extent to resolve. Finally, the panoramic laundry example shows successful selection of the requested row from a functional description. Together with the aggregate decline on implicit multi-target instructions, these cases motivate examining target completeness and mask extent alongside box localization.

\subsection{Ablation Study}
\label{sec:ablation}

The ablations attribute the gains to the two components introduced in Section \ref{sec:method}. The BBox-only variant is trained on the same Stage 2 grounding and reasoning data with the mask loss disabled, isolating pixel-level mask supervision. The Grounding-only variant keeps mask supervision but omits grounded reasoning data, isolating thinking with visual latents. Both tables include the original backbone as a reference.

\paragraph{Mask supervision and perception.}
Box supervision alone already improves both REC and perception over the backbone (Table \ref{tab:ablation_pixel_supervision}). The BBox-only variant gains 3.1, 4.9, and 2.2 validation REC points on RefCOCO, RefCOCO+, and RefCOCOg, and reaches 68.8 on MME-RW-en, 66.1 on MME-RW-cn, and 87.9 on CV-B, gains of 2.3, 5.8, and 2.5 points.

Adding mask supervision to this box-supervised setting raises all six metrics further. The full model's REC advantage over BBox-only ranges from 1.1 to 2.4 points, and its perception advantage from 0.2 to 1.4 points. The REC advantage is largest on RefCOCO+, at 2.4 points, and smallest on RefCOCO, at 1.1 points. Since RefCOCO+ excludes absolute location words, mask supervision helps most where the target must be separated from its neighbors by appearance and shape, which a box encodes only coarsely. Because the two variants differ only in the mask loss, these differences answer the question posed in Section \ref{sec:introduction}: supervising region extent in the MLLM's own representations improves box localization and fine-grained perception expressed in text.

\begin{table}[!tp]
    \centering
    \caption{Spatial supervision ablation on REC and general perception. Scores are validation-set Prec@0.5 and answer accuracy (\%), respectively. The first row is the Qwen3.5$_{\text{4B}}$ backbone~\citep{qwen3.5}.}
    \label{tab:ablation_pixel_supervision}
    \resizebox{\linewidth}{!}{
        \begin{tabular}{cccccccc}
            \toprule
            \multicolumn{2}{c}{\textbf{Stage 2 Supervision}} & \multicolumn{3}{c}{\textbf{Referring Expression Comprehension}} & \multicolumn{3}{c}{\textbf{General Perception}} \\
            \cmidrule(lr){1-2}
            \cmidrule(lr){3-5}
            \cmidrule(lr){6-8}
            BBox & Mask & RefCOCO & RefCOCO+ & RefCOCOg & MME-RW-en & MME-RW-cn & CV-B \\
            \midrule
            \ding{55} & \ding{55} & 90.4 & 84.8 & 88.0 & 66.5 & 60.3 & 85.4 \\
            \ding{51} & \ding{55} & 93.5 & 89.7 & 90.2 & 68.8 & 66.1 & 87.9 \\
            \rowcolor{Orange}
            \ding{51} & \ding{51} & 94.6 & 92.1 & 92.5 & 69.7 & 67.5 & 88.1 \\
            \bottomrule
        \end{tabular}
    }
\end{table}

\takeaway{1}{Mask supervision specifies target extent beyond what a box provides, helping most when appearance must disambiguate the target. Its gains on general perception benchmarks show that it improves the shared visual features.}

\paragraph{Grounded reasoning traces.}
Grounding-only training already yields strong referring performance (Table \ref{tab:ablation}). Its validation REC scores of 94.5, 91.5, and 92.0 exceed the backbone by 4.1, 6.7, and 4.0 points on RefCOCO, RefCOCO+, and RefCOCOg, respectively. Its RES scores are within 0.8 cIoU points of the full model on each dataset.

Building on this grounding-only setting, the full model adds grounded reasoning traces, which supervise the intermediate regions of thinking with visual latents (Section \ref{sec:thinking-visual-latents}). Relative to Grounding-only, its REC scores are 0.1 to 0.6 points higher, and its RES scores 0.5 to 0.8 points higher across the three validation splits. The gains concentrate in mask quality: the RES improvement matches or exceeds the REC improvement on every dataset. Mask supervision on intermediate regions in a CoT thus also benefits final-target prediction.

\begin{table}[!tp]
    \centering
    \caption{Training data ablation on REC and RES. Scores are validation-set Prec@0.5 and cIoU (\%), respectively. The first row is the Qwen3.5$_{\text{4B}}$ backbone~\citep{qwen3.5}; --- denotes unavailable results.}
    \label{tab:ablation}
    \resizebox{\linewidth}{!}{
        \begin{tabular}{cccccccc}
            \toprule
            \multicolumn{2}{c}{\textbf{Data}} & \multicolumn{3}{c}{\textbf{Referring Expression Comprehension}} & \multicolumn{3}{c}{\textbf{Referring Expression Segmentation}} \\
            \cmidrule(lr){1-2}
            \cmidrule(lr){3-5}
            \cmidrule(lr){6-8}
            Grounding & Reasoning & RefCOCO & RefCOCO+ & RefCOCOg & RefCOCO & RefCOCO+ & RefCOCOg \\
            \midrule
            \ding{55} & \ding{55} & 90.4 & 84.8 & 88.0 & --- & --- & --- \\
            \ding{51} & \ding{55} & 94.5 & 91.5 & 92.0 & 82.7 & 79.1 & 80.6 \\
            \rowcolor{Orange}
            \ding{51} & \ding{51} & 94.6 & 92.1 & 92.5 & 83.4 & 79.9 & 81.1 \\
            \bottomrule
        \end{tabular}
    }
\end{table}

\takeaway{2}{Grounding data drives most of the referring gain, while grounded reasoning traces mainly sharpen final masks. Supervising intermediate regions thus refines region extent.}

\FloatBarrier
\section{Related Work}
\label{sec:related-work}

\subsection*{Visual Grounding in MLLMs}

MLLMs commonly ground language by generating box coordinates in the text stream, as in Shikra, Ferret, and CogVLM~\citep{chen2023shikra,zhang2024ferret,wang2024cogvlm,yu2026groundingpi,yu2026groundanything}; Florence-2 casts grounding and other vision tasks as sequence prediction~\citep{xiao2024florence}. Under this interface, spatial supervision reaches the model only through coordinate tokens, and the extent of the referenced region remains implicit. Youtu-VL extends autoregressive supervision to visual targets alongside language and examines how such supervision shapes multimodal learning~\citep{wei2026youtu}. For pixel-level grounding, RES makes region extent explicit by requiring a pixel-level mask for the referenced target.

\subsection*{Segmentation with MLLMs}

Most MLLM-based segmentation systems obtain spatial features from a dedicated segmenter. LISA conditions a SAM-based branch on a generated token~\citep{lai2024lisa,kirillov2023segment}; LISA++ adds instance-aware outputs, and GSVA handles multiple or empty targets~\citep{yang2023lisa++,xia2024gsva}. GLaMM and OMG-LLaVA interleave masks with grounded conversation~\citep{rasheed2024glamm,zhang2024omgllava}, and Sa2VA couples an MLLM with SAM 2 for image and video tasks~\citep{yuan2025sa2va,ravi2024sam2}. Seg-Zero trains a reasoning model to prompt a frozen segmenter~\citep{liu2025segzero}, and InstructSAM learns parallel instance queries that condition SAM 3~\citep{yuan2026instructsam,carion2026sam3}. Qwen3-VL-Seg replaces the segmenter with a lightweight supplementary encoder and combines multi-scale feature fusion with box-guided decoding and iterative mask refinement~\citep{yao2026qwen3vlseg}.

Alongside these dedicated segmentation branches, another line derives spatial predictions from the MLLM's own representations or output interface. PixelLM decodes masks from features at multiple layers of its vision encoder~\citep{ren2024pixellm}, and MLLMSeg decodes them from shallow vision-encoder features fused with the language model's visual tokens~\citep{wang2025mllmseg}; both keep the vision encoder frozen. ClawMachine refers to entities through groups of visual tokens, and PaDT maps patch-derived reference tokens to boxes or masks with a lightweight decoder~\citep{ma2024clawmachine,su2025patch}. Text4Seg represents segmentation through textual descriptors, while UFO combines a language interface with embedding retrieval~\citep{lan2025text4seg,tang2025ufo}. Both lines evaluate the benefit of mask supervision mainly on segmentation and grounding. To study the effect beyond spatial prediction, \REPORTNAME decodes masks from a single patch-level grid of the shared vision encoder and evaluates whether the resulting supervision also improves fine-grained perception, as measured by text answers.

\subsection*{Grounded and Latent Visual Reasoning}

Fine-grained visual reasoning often requires details beyond a global image description. To capture such details, Cambrian-1 integrates high-resolution features through a spatially aware connector~\citep{tong2024cambrian}; V* searches for relevant regions, and Divide, Conquer, and Combine processes high-resolution patches with textual summaries~\citep{wu2023vstar,wang2024divide}. OpenAI o3 thinks with images through cropping, zooming, and other image transformations~\citep{openai2025thinkingimages}, and DeepEyes, Pixel Reasoner, Chain-of-Focus, and Mini-o3 learn or control such image interactions during visual search~\citep{zheng2025deepeyes,wang2025pixelreasoner,zhang2025chainoffocus,lai2025minio3}.

Complementing such image interactions, other work either grounds intermediate reasoning steps explicitly or carries them in visual form. Visual CoT supervises intermediate box references~\citep{shao2024visualcot}, and UniPixel links intermediate mask pointers to object-centric reasoning~\citep{liu2025unipixel}. Meanwhile, Mirage interleaves latent visual tokens with text during multimodal reasoning~\citep{yang2025mirage}, and LVR performs reasoning in the visual embedding space~\citep{li2026lvr}. Thinking with visual latents connects these directions: each region reference produces a visual latent that the shared mask decoder maps to a mask, and the mask loss on that latent updates the backbone. Intermediate regions thus receive pixel-level supervision and remain inspectable through the decoded masks.

\section{Conclusion}
\label{sec:conclusion}

We presented \REPORTNAME, a 4B MLLM that decodes region masks from the patch-level grid of its shared vision encoder and from visual latents, the language-model states of generated mask tokens. The mask loss therefore reaches the representations used for language generation, and thinking with visual latents extends this pixel-level supervision to the regions named in a grounded CoT.

Without an additional spatial encoder, \REPORTNAME achieves the highest cIoU on all eight RES splits, improves REC over its Qwen3.5 backbone on every split, and achieves the highest threshold-averaged F1 for multi-instance grounding on COCO 2017. Moreover, it surpasses the backbone on every fine-grained perception benchmark; the BBox-only ablation attributes a further gain on all evaluated REC and perception metrics to mask supervision, and grounded reasoning further raises REC and RES scores. On LensBench, it leads in cIoU on aerial, driving, and panoramic imagery, although implicit instructions remain harder, especially for multiple targets in panoramic scenes.

\clearpage
\bibliographystyle{plainnat}
\bibliography{ref}

\clearpage
\phantomsection
\section*{Contributors}
\label{contributors}

\REPORTNAME was developed by the SeerRay Team. Core contributors, contributors, and supervisors are listed below. Names are ordered alphabetically by surname within each group, without indicating the relative extent of individual contributions. Corresponding authors are marked with \textsuperscript{\dag}.

\noindent
\begin{minipage}[t]{0.45\textwidth}
\textbf{\textsf{\xiaomiblue{Core Contributors}}}
\begin{itemize}[leftmargin=12pt, itemsep=1pt, parsep=0pt, topsep=4pt]
  \item Pei Fu\textsuperscript{\dag}
  \item Yike Liu
  \item Yue Sun
  \item Hanmo Wang
  \item Xiuwen Xi
  \item Ruoceng Zhang
  \item Shaojie Zhang
\end{itemize}
\end{minipage}%
\hfill
\begin{minipage}[t]{0.45\textwidth}
\textbf{\textsf{\xiaomiblue{Contributors}}}
\begin{itemize}[leftmargin=12pt, itemsep=1pt, parsep=0pt, topsep=4pt]
  \item Tingwei Huang
  \item Chao Jiang
  \item Hang Li
  \item Bingrui Liu
  \item Zhenbo Luo
  \item Heng Qu
  \item Changjian Xu
  \item Ming Xu
\end{itemize}
\end{minipage}

\vspace{8pt}
\noindent
\textbf{\textsf{\xiaomiblue{Supervisors}}}
\begin{itemize}[leftmargin=12pt, itemsep=1pt, parsep=0pt, topsep=4pt]
  \item Jian Luan\textsuperscript{\dag}
\end{itemize}

\begin{NoHyper}\let\thefootnote\relax\footnotetext{\textsuperscript{\dag} Corresponding authors: Pei Fu (\texttt{fupei1@xiaomi.com}) and Jian Luan (\texttt{luanjian@xiaomi.com}).}\end{NoHyper}

\clearpage
\phantomsection
\addcontentsline{toc}{section}{Appendix}
\label{sec:appendix}
\addtocontents{toc}{\protect\setcounter{tocdepth}{-1}}
\beginappendix
\appendix

\FloatBarrier

\section{Conversation Format and Examples}
\label{sec:conversation-examples}

The examples below illustrate how spatial predictions are serialized in a response. A \texttt{\textless{}bbox\textgreater{}} block contains box coordinates, and a mask block contains the special \texttt{\textless{}|mask|\textgreater{}} token. The contextual state at this token is the visual latent passed to the decoder in Section \ref{sec:model-architecture}.

\paragraph{Explicit Instruction: Box Output}
The clothing description directly identifies the requested person. The response returns a single box after an empty reasoning block.

\begin{Verbatim}[fontsize=\small,baselinestretch=0.95,frame=single,framesep=1.5mm,breaklines=true,breakautoindent=true,breakindent=1em,breaksymbolleft={},commandchars=\@\(\)]
[
    {
        "role": "user",
        "content": "<image>\nLocate all instances in the image that match the description \"@textbf(the lady with the blue shirt)\". Report bbox coordinates."
    },
    {
        "role": "assistant",
        "content": "<think>\n\n</think>\n\n<bbox>[244, 469, 557, 745]</bbox>"
    }
]
\end{Verbatim}

\paragraph{Implicit Instruction: Box and Mask Outputs}
The instruction specifies a protective function that requires the model to infer the target category. The response identifies and describes the selected helmet, then returns its box and a mask token whose visual latent represents that target.

\begin{Verbatim}[fontsize=\small,baselinestretch=0.95,frame=single,framesep=1.5mm,breaklines=true,breakautoindent=true,breakindent=1em,breaksymbolleft={},commandchars=\@\(\)]
[
    {
        "role": "user",
        "content": "<image>\nLocate all instances in the image that match the description \"@textbf(If we want to protect our heads when participating in contact sports like football or baseball, what object in the picture should we wear?)\". Report bbox coordinates and masks."
    },
    {
        "role": "assistant",
        "content": "<think>\nTo protect the head in contact sports, the appropriate item to wear is a helmet. In the center of the image, sitting on a wooden shelf, is a glossy blue helmet featuring the number 99 and an Oilers logo. This object is clearly a piece of protective headgear. While there are other padded items visible on the left and right edges, the blue helmet in the middle is the most distinct and complete example of the required safety equipment. Therefore, this central blue region is the correct target to identify.\n</think>\n\n<bbox>[190, 169, 756, 924]</bbox><mask><|mask|></mask>"
    }
]
\end{Verbatim}

\clearpage
\section{Qualitative Analysis}
\label{sec:case-study}

The qualitative examples connect the quantitative results in Section \ref{sec:experiments} to target selection, region extent, and intermediate region references. Grounding examples show the boxes and masks predicted for requested targets. Complementing these grounding examples, VQA examples show how thinking with visual latents makes intermediate regions inspectable: each region discussed in the reasoning trace is linked to its decoded mask and to the final answer. For these VQA visualizations, we replay the stored completion through the same model to obtain the contextual states used for mask decoding.

\begin{figure}[!htbp]
    \centering
    \includegraphics[width=\textwidth]{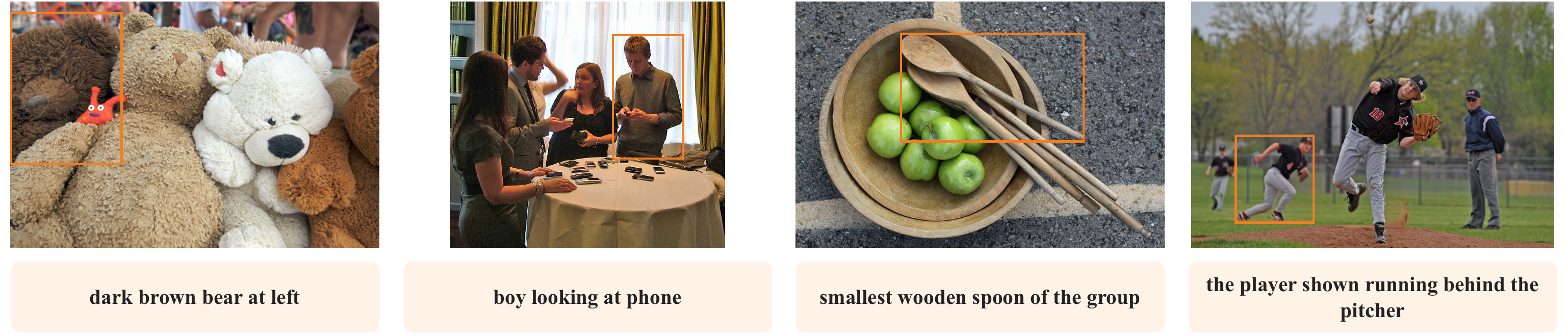}
    \caption{Box predictions for referring expressions based on appearance, action, relative size, and spatial relations. Each orange box marks the instance selected by the model.}
    \label{fig:rec-cases}
\end{figure}

\begin{figure}[!htbp]
    \centering
    \includegraphics[width=\textwidth]{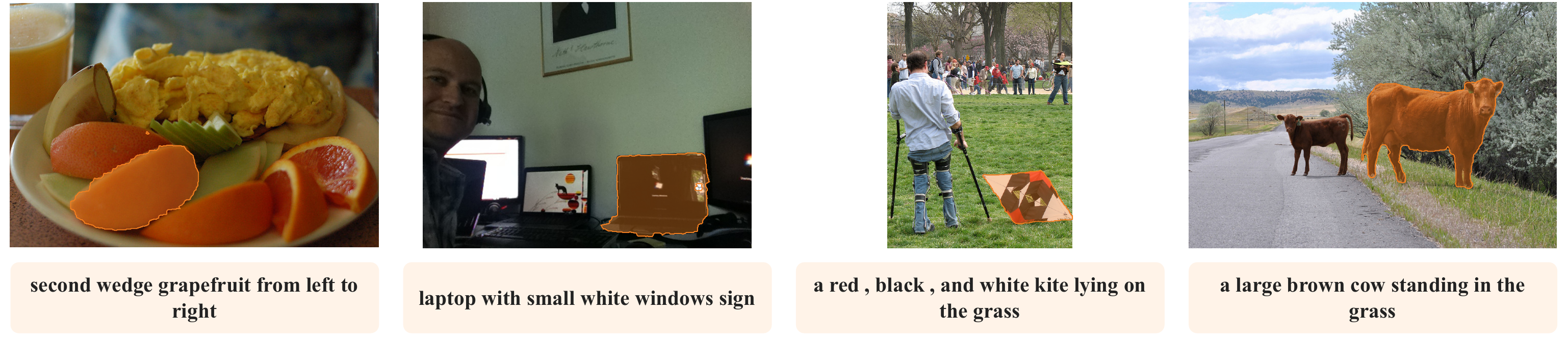}
    \caption{Mask predictions for targets identified by referring expressions. Orange overlays delineate the selected fruit wedge, laptop, kite, and cow. The accompanying expressions specify the intended referent in each scene.}
    \label{fig:res-cases}
\end{figure}

\begin{figure}[!htbp]
    \centering
    \includegraphics[width=\textwidth]{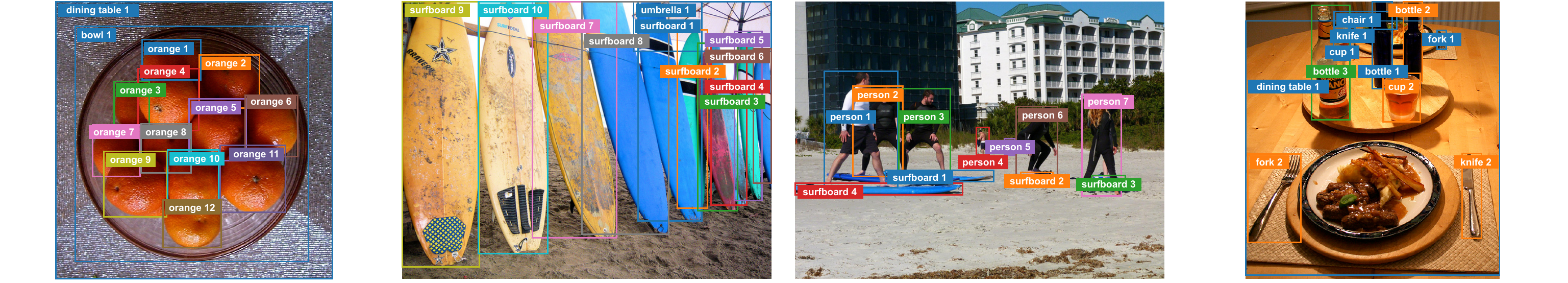}
    \caption{Multi-object box predictions in scenes with repeated instances and multiple categories. Labels identify the predicted category and distinguish instances within each category.}
    \label{fig:coco-cases}
\end{figure}

\paragraph{Referring Expression Grounding and Multi-Object Localization}
Figures \ref{fig:rec-cases} and \ref{fig:res-cases} illustrate target selection and region delineation from appearance, action, relative size, and spatial relations. The masks distinguish requested instances from nearby objects, including a particular fruit wedge and a cow beside another cow. Figure \ref{fig:coco-cases} shows multi-instance predictions, where individual boxes distinguish repeated or overlapping objects such as oranges in a bowl and surfboards in a rack.

\begin{figure}[!htbp]
    \centering
    \includegraphics[width=\textwidth,height=0.72\textheight,keepaspectratio]{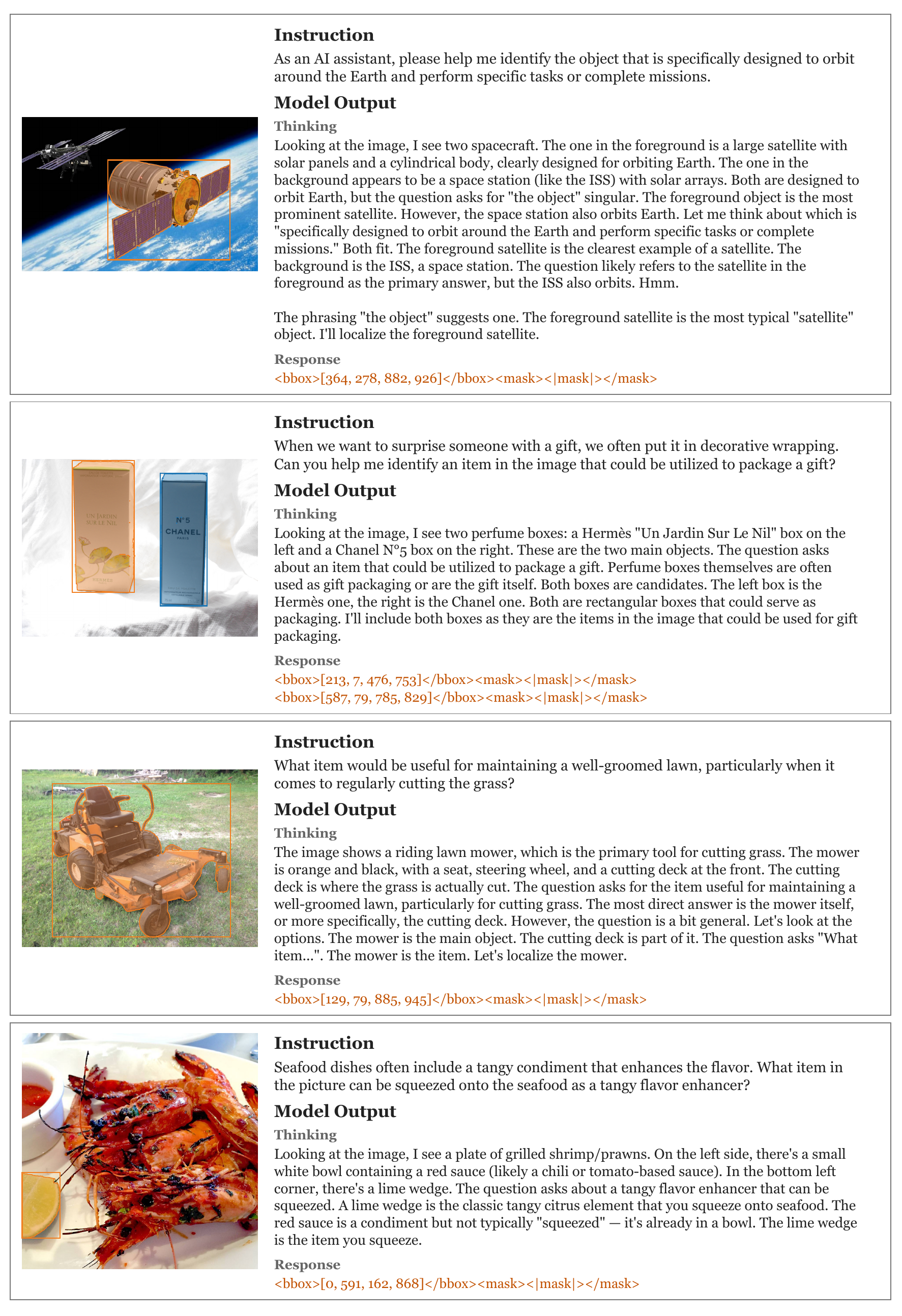}
    \caption{Predictions from functional descriptions. Each example pairs the instruction and response with the predicted boxes and masks in orange. The spacecraft and packaging traces explicitly consider multiple candidate referents.}
    \label{fig:reasonseg-cases}
\end{figure}

\clearpage
\paragraph{Grounding Functional Descriptions}
Figure \ref{fig:reasonseg-cases} illustrates target selection from functional descriptions, including objects used to cut grass or add citrus flavor to seafood. The responses connect each requested function to visible candidates. For example, the spacecraft trace considers two candidates before selecting the foreground object, while the gift-packaging example returns both boxes.

\begin{figure}[!htbp]
    \centering
    \includegraphics[width=\textwidth,height=0.72\textheight,keepaspectratio]{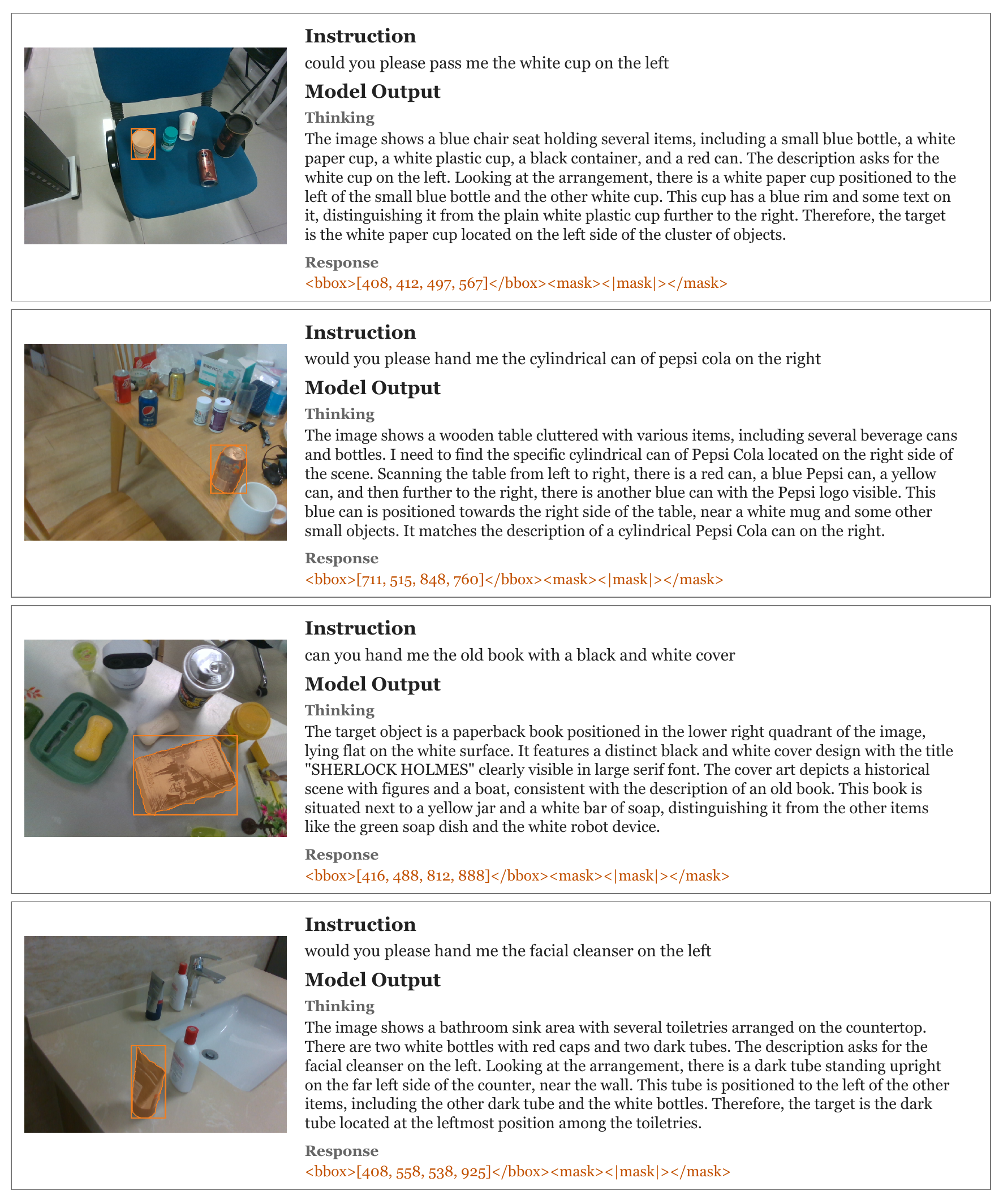}
    \caption{Object selection from requests that combine category, appearance, and position. Orange boxes and masks show the predictions alongside the reasoning and response.}
    \label{fig:roborefit-cases}
\end{figure}

\clearpage
\paragraph{Object Selection in Cluttered Scenes}
Figure \ref{fig:roborefit-cases} shows referring requests that combine object identity with attributes or relative position in cluttered scenes. These cues select a cup, a beverage can, a book, and a cleanser among nearby objects. Specifically, positional phrases distinguish the requested cup and cleanser from other instances of the same category, and the masks delineate the selected objects.

\begin{figure}[!htbp]
    \centering
    \includegraphics[width=\textwidth,height=0.72\textheight,keepaspectratio]{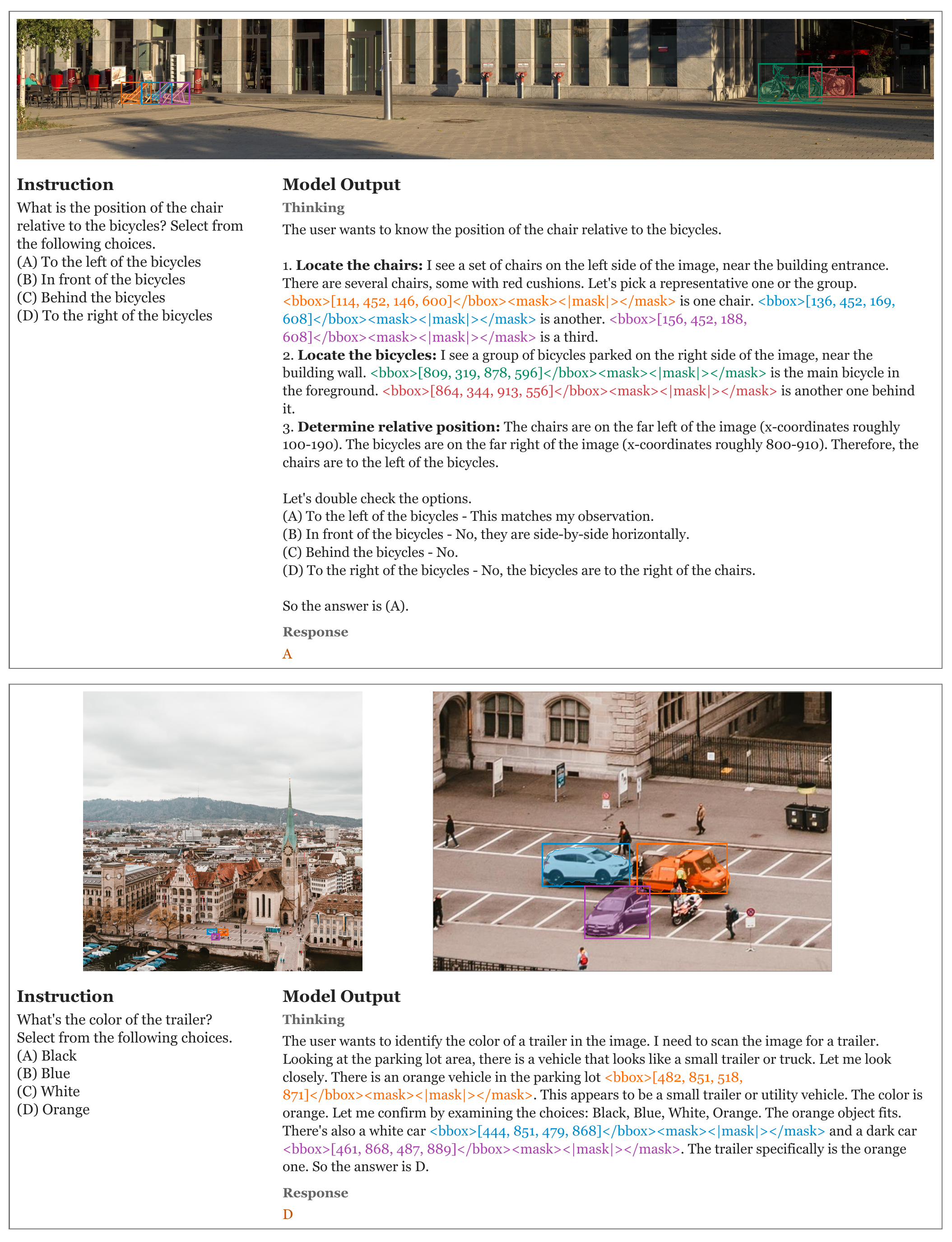}
    \caption{Region references in spatial comparison (top) and vehicle-color recognition (bottom). Matching colors link the response tokens to image regions. The enlarged view shows the small targets in the second example.}
    \label{fig:hrbench-cases}
\end{figure}

\clearpage
\paragraph{Spatial Comparison and Color Recognition}
Figure \ref{fig:hrbench-cases} illustrates visual latents associated with spatial comparison and color recognition. The first trace localizes chairs and bicycles before comparing their horizontal positions and selecting answer A. The second identifies a small vehicle in the wider scene and selects answer D, orange. Matching colors connect each region reference to its box and decoded mask, and an enlarged view shows the small target.

\begin{figure}[!htbp]
    \centering
    \includegraphics[width=\textwidth,height=0.72\textheight,keepaspectratio]{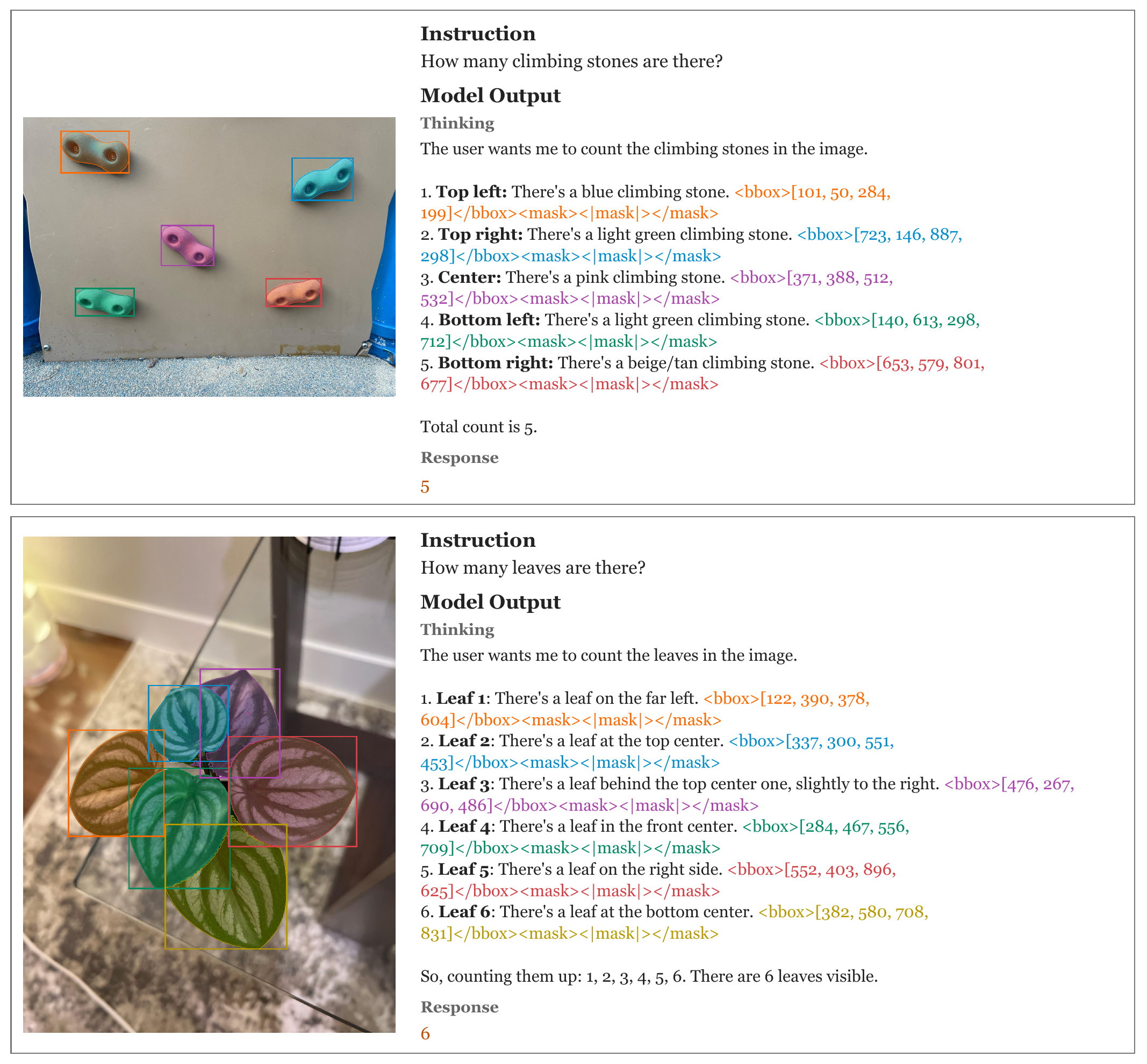}
    \caption{Instance-by-instance counting of climbing stones (top) and leaves (bottom). The traces contain five and six region references, respectively. Each colored token block corresponds to a box and a mask.}
    \label{fig:countqa-cases}
\end{figure}

\clearpage
\paragraph{Counting with Instance References}
Figure \ref{fig:countqa-cases} illustrates instance-by-instance counting with visual latents. The traces enumerate five spatially separated climbing stones and six leaves, several of which overlap, and then return the corresponding counts. Matching colors link each enumerated item to its predicted box and mask, so the region behind each counting step can be inspected.

\FloatBarrier

\addtocontents{toc}{\protect\setcounter{tocdepth}{2}}

\end{document}